\documentclass{article}

\usepackage[main, final]{neurips_2026}
\usepackage{tcolorbox}
\tcbuselibrary{listings, breakable}

\usepackage[utf8]{inputenc} 
\usepackage[T1]{fontenc}    
\usepackage{hyperref}       
\usepackage{url}            
\usepackage{booktabs}       
\usepackage{amsfonts}       
\usepackage{nicefrac}       
\usepackage{microtype}      
\usepackage{xcolor}         
\usepackage{amsmath}
\usepackage{multirow}
\usepackage{subcaption}
\usepackage{tabularx}
\usepackage{graphicx}
\usepackage{wrapfig}

\newcommand\blfootnote[1]{%
  \begingroup
  \renewcommand\thefootnote{}\footnote{#1}%
  \addtocounter{footnote}{-1}%
  \endgroup
}

\title{Neural networks are more modular than single neurons suggest}

\author{%
  Shreya Saha$^{1}$ \quad
  Benjamin Bergen$^{2}$ \quad
  Meenalshi Khosla$^{2,3}$ \\[2mm]
  $^{1}$Electrical and Computer Engineering \quad
  $^{2}$Cognitive Science Department\quad
  $^{3}$Computer Science and Engineering \\
  University of California, San Diego
}

\begin{document}

\maketitle

\begin{abstract}
Neural networks routinely develop representations in which individual neurons contribute to multiple distinct functions, leading to the widespread view that such networks lack modular organization. We challenge this conclusion by introducing a formally nested hierarchy of modularity: unit-level, which requires distinct computations to occupy disjoint neurons; orthogonal, which asks whether a rotation reveals functionally independent subspaces; and linear, which asks whether a bounded linear map can separate task-relevant computations into independent subspaces. We develop a unified framework for learning decompositions at each level and evaluating them causally through selective ablation. Across vision models, language models and multi-task Recurrent Neural Networks (RNNs), we find that networks which lack unit-level modularity nonetheless often exhibit clear functional modularity at the subspace level. In vision and language models, content versus style and syntax versus semantics respectively dissociate under subspace ablation despite sharing overlapping populations of neurons. These learned subspaces 
also align differentially with brain data, with content and style subspaces predicting higher and lower visual cortex respectively, and semantic subspaces outperforming syntactic 
ones in predicting language cortex activity. In RNNs trained on multiple cognitive tasks and previously shown to lack modular structure, our decomposition recovers sharp double dissociations between tasks. The absence of unit-level modularity frequently coexists with clear subspace-level modularity across domains,  enabling targeted intervention on specific computations even in networks where no individual neuron is specialized, with broad implications for mechanistic interpretability and for understanding functional organization in both biological and artificial neural systems.\blfootnote{All code is publicly available at: \href{https://github.com/NeuroML-Lab/Disentangling-representations}{https://github.com/NeuroML-Lab/Disentangling-representations}}
  
\end{abstract}

\section{Introduction}

The idea that complex systems decompose into functionally specialized components, often known as \emph{modularity}, is among the oldest organizing principles in neuroscience and a central question for understanding artificial neural networks \cite{jacobs1991adaptive, zeki1978functional, fodor1983modularity, kanwisher2010functional}. In neuroscience, the gold standard for establishing modularity is double dissociation: damage to one region selectively impairs one function while sparing another, and damage to a second region produces the reverse pattern\citep{moscovitch1997special, damasio1982prosopagnosia}. This establishes that two functions rely on separable neural machinery, not merely that they activate different populations. 

In practice, however, modularity is  assessed not through direct causal intervention but through a proxy, like  \emph{selectivity}: whether individual neurons or voxels respond preferentially to one function over another. This proxy has yielded clear positive results in some domains \cite{kanwisher1997fusiform} but notable null results in others. For example, extensive investigation of the human language network has failed to find reliable separation for syntactic versus semantic processing at the level of individual voxels \citep{fedorenko2020lack}, yet whether these functions might dissociate at a different level of description remains an open question.

\begin{figure}[h!]
\centering
\includegraphics[width=0.9\linewidth]{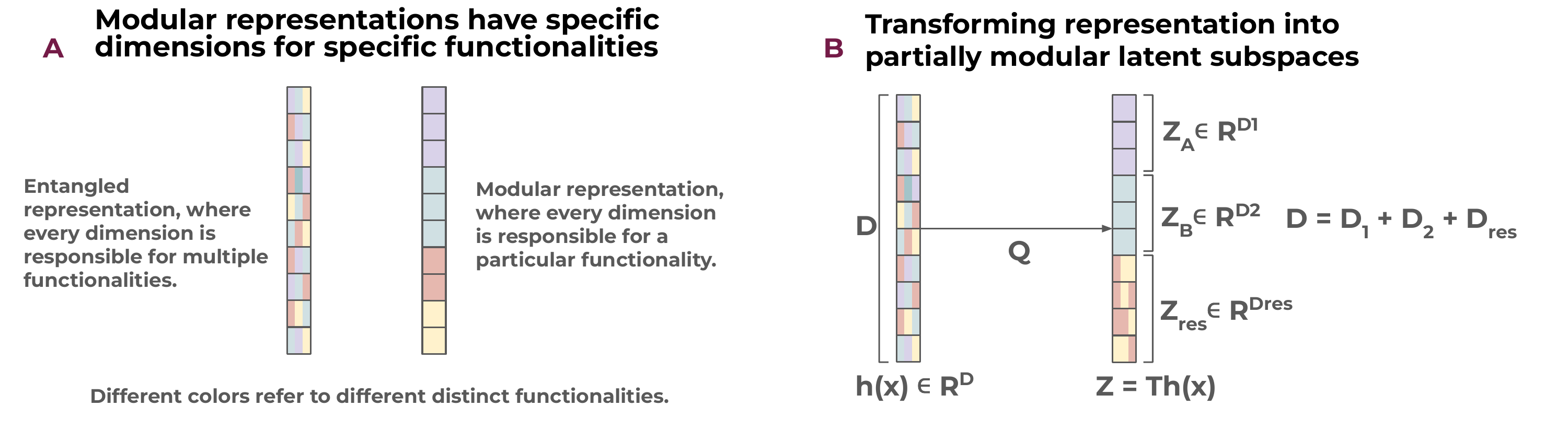}
\caption{\textbf{Illustration of functional modularity and our proposed decomposition framework.}
(A) In an entangled representation, individual dimensions participate in multiple functionalities, such that the representations of different functions overlap across dimensions. In an ideally modular representation, each functionality occupies a distinct, non-overlapping set of dimensions, allowing the functions to be independently manipulated.
(B) Our framework applies an information-preserving transformation to an entangled representation to separate two target functionalities. Ideally, the transformed representation decomposes into three subspaces: two function-specific subspaces corresponding to the target functionalities being dissociated, and a residual subspace containing information associated with other, non-target functionalities.}
\label{fig:modularity_schematic}
\end{figure}

In artificial neural networks, modularity has been pursued along two distinct lines. One line seeks to build modular structure into the architecture itself, using explicit routing mechanisms, attention-based module selection, or other inductive biases that encourage computational specialization \citep{goyal2022inductive, csordas2020neural}. A second line asks whether modularity emerges spontaneously in networks that lack such inductive biases. Here, the dominant approach mirrors the neuroscience tradition: examine individual neurons for selectivity. A network is considered modular if distinct tasks are carried by disjoint sets of selective neurons. By this criterion, modularity has been demonstrated in networks with element-wise nonlinearities, where clustering of task variance across units reveals groups of functionally specialized neurons \citep{yang2019task}. But the same criterion frequently returns a negative verdict. Networks with smooth, saturating nonlinearities like tanh, as well as most transformer-based models, exhibit pervasive mixed selectivity: individual units respond to combinations of features (often known as superposition) or task variables with no clear specialization \citep{driscoll2024flexible, rigotti2013importance}. 

We argue that this conclusion conflates two distinct claims: that individual neurons are not specialized, and that the network's computation is not modular. The first may be true while the second is false. A system can organize functionally independent computations into separable subspaces of the population activity without any single neuron being dedicated to a single function (Figure \ref{fig:modularity_schematic} A). Mixed selectivity at the unit level is perfectly compatible with modular organization at the subspace level - the modules are simply rotated off the coordinate axes.

To make this argument precise, we introduce a hierarchy of three formally nested notions of modularity. \textbf{Unit-level modularity} is the classical notion: distinct computations occupy disjoint sets of neurons, and ablating each set selectively impairs one function but not the other. \textbf{Orthogonal modularity} relaxes this by asking whether there exists a rotation of the representation space that reveals functionally independent subspaces. \textbf{Linear modularity} relaxes further, asking whether a bounded invertible linear transformation can separate computations into independent subspaces, permitting the bases to be non-orthogonal.

We evaluate whether a candidate decomposition constitutes genuine functional modularity (rather than a statistical artifact) through ablation. We say a decomposition is modular if ablating each of two subspaces selectively impairs one unique computation while preserving the other (a double dissociation). We develop a unified optimization framework (Figure \ref{fig:modularity_schematic} B) for learning decompositions at each level that are optimized for task sufficiency, and evaluate them through interventional ablation on held-out data. We then evaluate the strength of functional separation, accounting for both selective impairment and collateral damage.

Our notion of modularity is distinct from \emph{distributed coding} and \emph{superposition} \cite{elhage2022toy, henighan2023superposition}, which describe how features are represented within a population. Distributed coding asks whether a feature is represented across many neurons, whereas superposition asks whether multiple features share a limited set of representational dimensions. Neither concept directly addresses whether different computations are functionally separable. A representation can be distributed while remaining functionally modular: for example, a dense rotation can distribute two functionally independent computations across every neuron while preserving their distinct subspaces. Conversely, a distributed representation need not contain subspaces that selectively support different computations. This also distinguishes our question from sparse autoencoders and related feature discovery methods \cite{bricken2023towards, templeton2026scaling, huben2024sparse}, which seek a new basis of more interpretable or monosemantic features. Our goal is not to create monosemantic neurons or recover individual features, but to determine whether the existing representation already contains task-selective functional subspaces. Crucially, neither distributed coding nor superposition is defined by, or established through, a double dissociation. They characterize how features are distributed across and share representational coordinates; they do not determine whether different computations are independently supported. Functional modularity asks a distinct causal question: \emph{can one computation be selectively disrupted while another is spared, and vice versa?}

We apply this optimization framework across 3 domains. In multitask recurrent neural networks trained on cognitive tasks from \citet{yang2019task}, we revisit the well-established finding that networks with certain activation functions lack the clustered unit structure found in networks with rectifying activations. Our subspace decomposition recovers strong double dissociations in these networks that are invisible to unit-level analysis, confirming the hypothesis of \citet{driscoll2024flexible}. In vision models, we demonstrate that content and style occupy separable linear subspaces despite being encoded by overlapping neural populations. In language models, we show that syntactic and semantic processing similarly dissociate at the subspace level even though the same units participate in both. We further show that these decomposed vision and language subspaces exhibit brain alignment with established functional organization in both visual and language cortex. Across all three domains, we find a consistent pattern: an absence of unit-level modularity paired with clear subspace-level modularity. These results demonstrate that modularity cannot be assessed at a single level of description, and that the hierarchy we introduce provides a principled framework for determining when and at what level of analysis functional specialization exists in both biological and artificial networks.

\section{Methods}
\label{subsec:methods}

\subsection{A hierarchy of linear decompositions}
Let $\mathbf{h}(x) \in \mathbb{R}^D$ denote the representation of an input $x$ 
at a given layer of a neural network. We seek a linear transformation 
$Q \in \mathbb{R}^{D \times D}$ such that the transformed representation 
$\mathbf{z} = Q\mathbf{h}(x)$ decomposes into non-overlapping coordinate blocks (Figure \ref{fig:modularity_schematic} B):
\begin{equation}
    \mathbf{z} = \big[\,\mathbf{z}_A \in \mathbb{R}^{d_A},\;\; 
    \mathbf{z}_B \in \mathbb{R}^{d_B},\;\; 
    \mathbf{z}_{\mathrm{res}} \in \mathbb{R}^{d_r}\,\big], 
    \qquad d_A + d_B + d_r = D.
    \label{eq:partition}
\end{equation}
Subspace $\mathbf{z}_A$ is intended to capture computation relevant to 
function $A$, subspace $\mathbf{z}_B$ to function $B$, and 
$\mathbf{z}_{\mathrm{res}}$ captures all remaining structure. The residual 
subspace is essential: any representation encodes information beyond two 
targeted functions, and discarding it would conflate compression with 
decomposition.

We study three families of transformations, ordered from most to least 
constrained (More details can be found in Appendix section \ref{subsec:constraints}):

\textbf{Permutation:} $Q$ reorders coordinates without mixing 
    them, testing whether the representation is already axis-aligned with 
    respect to the target functions.

\textbf{Orthogonal:} $Q^\top Q = I$, allowing rigid rotations 
    that preserve all inner products and norms.

\textbf{Invertible:} $Q$ is a general invertible matrix with 
    bounded condition number $\kappa(Q) \leq \kappa_{\max}$, permitting 
    non-orthogonal bases.
This nesting, permutation $\subset$ orthogonal $\subset$ invertible,
forms a diagnostic hierarchy. The most constrained family under which 
functional separation succeeds reveals how computations are geometrically 
organized in the original representation.

\subsection{Training objectives}
The decomposition $Q$ must be trained with a supervisory signal that 
encourages $\mathbf{z}_A$ and $\mathbf{z}_B$ to capture their respective 
target functions. We employ two complementary objectives depending on the 
domain.

\paragraph{Task-sufficiency objective (RNNs)}
\label{sec:methods_rnn}
For recurrent networks trained on explicit tasks with known input--output mappings, we optimize $Q$ so that each subspace is \emph{sufficient} 
for its corresponding task. Given hidden states $\{h^A_i\}$ from task $A$ trials with targets $\{y^A_i\}$, and similarly for task $B$, we train 
linear readout heads $f_A$ and $f_B$ jointly with $Q$ to minimize: $\mathcal{L}_{\mathrm{task}} = 
    \mathcal{L}(f_A(\mathbf{z}_A),\, y^A) + 
    \mathcal{L}(f_B(\mathbf{z}_B),\, y^B) + 
    \lambda\, \mathcal{L}_{\mathrm{reg}}$,
where $\mathcal{L}$ is the task loss (MSE for Yang, cross-entropy for 
Mante) and $\mathcal{L}_{\mathrm{reg}}$ is a level-specific regularizer 
(Sinkhorn entropy for permutations, Cayley parameterization for orthogonal, 
condition number penalty for invertible). This objective directly optimizes for functional relevance: each subspace must contain enough information to perform its task. Please refer to Appendix section \ref{app:rnnnetworkappendix} for detailed descriptions regarding the experiment details and Appendix section \ref{subsec:datasets} for dataset description.

\paragraph{Contrastive objective (vision and language models)}
\label{sec:methods_contrastive}

For pretrained models without paired task labels, we use contrastive learning on \emph{minimal pairs}: input pairs that vary along exactly one axis of interest while holding the other fixed. For each anchor input $x$, we construct a subspace-$A$ counterpart $x^+_A$ (sharing structure along $A$, differing along $B$) and a subspace-$B$ counterpart 
$x^+_B$ (the reverse). We train $Q$ with two contrastive losses~\cite{khosla2020supervised},
one per subspace: 

\begin{equation}
    \mathcal{L}_{A} = 
    -\frac{1}{N}\sum_{i=1}^{N} \log 
    \frac{
        \exp\!\big(\mathrm{sim}(\mathbf{z}_{A}(x_i),\, 
        \mathbf{z}_{A}(x^+_{A,i})) / \tau\big)
    }{
        \sum_{j \neq i} 
        \exp\!\big(\mathrm{sim}(\mathbf{z}_{A}(x_i),\, 
        \mathbf{z}_{A}(x^-_j)) / \tau\big)
    },
    \label{eq:loss_contrastive}
\end{equation}
and symmetrically for $\mathcal{L}_B$. The total objective is 
$\mathcal{L} = \mathcal{L}_A + \mathcal{L}_B$. Because the two losses act on non-overlapping coordinate blocks of $\mathbf{z}$, gradients from 
each loss update disjoint columns of $Q$, ensuring geometric separation by construction. More details on constraining $Q$ to be invertible, orthogonal or permutation based can be found in Appendix section \ref{subsec:constraints}.

For \textbf{content--style} decomposition in vision models, $x^+_{\mathrm{content}}$ is a different stylization of the same scene and $x^+_{\mathrm{style}}$ is a different scene rendered in the same artistic style. For \textbf{syntax--semantics} decomposition in language models, $x^+_{\mathrm{semantics}}$ is a paraphrase preserving meaning and $x^+_{\mathrm{syntax}}$ is a sentence sharing the same grammatical 
template with different lexical content.

More experimental details can be found in Appendix section \ref{subsec:exptdetails}, and more details on datasets used for the above decompositions can be found in Appendix section \ref{subsec:datasets}.

\subsection{Evaluation}
\label{sec:methods_eval}

We evaluate decompositions differently depending on whether the 
domain admits direct causal intervention. Detailed description of the datasets is in Appendix section \ref{subsec:datasets}.

\textbf{Causal ablation (RNNs).}
For recurrent networks with explicit task structure, we zero out subspace $\mathbf{z}_A$ (or $\mathbf{z}_B$) at every timestep during inference by applying $Q$, setting the relevant coordinates to zero, and mapping back via $Q^{-1}$. We then measure task accuracy under each ablation condition. A decomposition exhibits functional modularity if ablating $\mathbf{z}_A$ selectively impairs function $A$ while sparing function $B$, and vice versa, i.e., a double dissociation (see Eq.~\ref{eq:ddi}). Complete details on training and inference on Continual recurrent neural networks can be found in Appendix section \ref{app:rnnnetworkappendix}.

\textbf{Retrieval-based evaluation (vision and language).}
For pretrained vision and language models, where task performance is not defined by a single scalar output, we evaluate subspace quality through nearest-neighbor retrieval. For each query input, we retrieve the top-$k$ nearest neighbors in each subspace and measure the fraction that share the correct property with the query. In vision models, a well-disentangled decomposition retrieves content-matched neighbors in the content subspace and style-matched neighbors in the style subspace, with minimal cross-contamination. In language models, we pair each sentence with one semantic counterpart (a paraphrase preserving meaning) and multiple syntactic counterparts (sentences sharing the same 
grammatical template with different lexical content), and measure whether each subspace preferentially retrieves its intended counterpart type. Full metrics are defined in Sections~\ref{sec:vision_results} and~\ref{sec:language_results}. Full details on the vision and language models used is present in Appendix section \ref{subsec:model_details}, and unless otherwise mentioned we use the penultimate layer representation (average pooled in case of vision models and average across all tokens in case of language models).

\section{Results}
\label{subsec:results}

\subsection{Multitask cognitive RNNs (Yang's task)}
\label{sec:rnn_results}

\begin{figure}[t]
    \centering
\includegraphics[
    width=0.88\textwidth,
    trim=0 130 0 0,
    clip
]{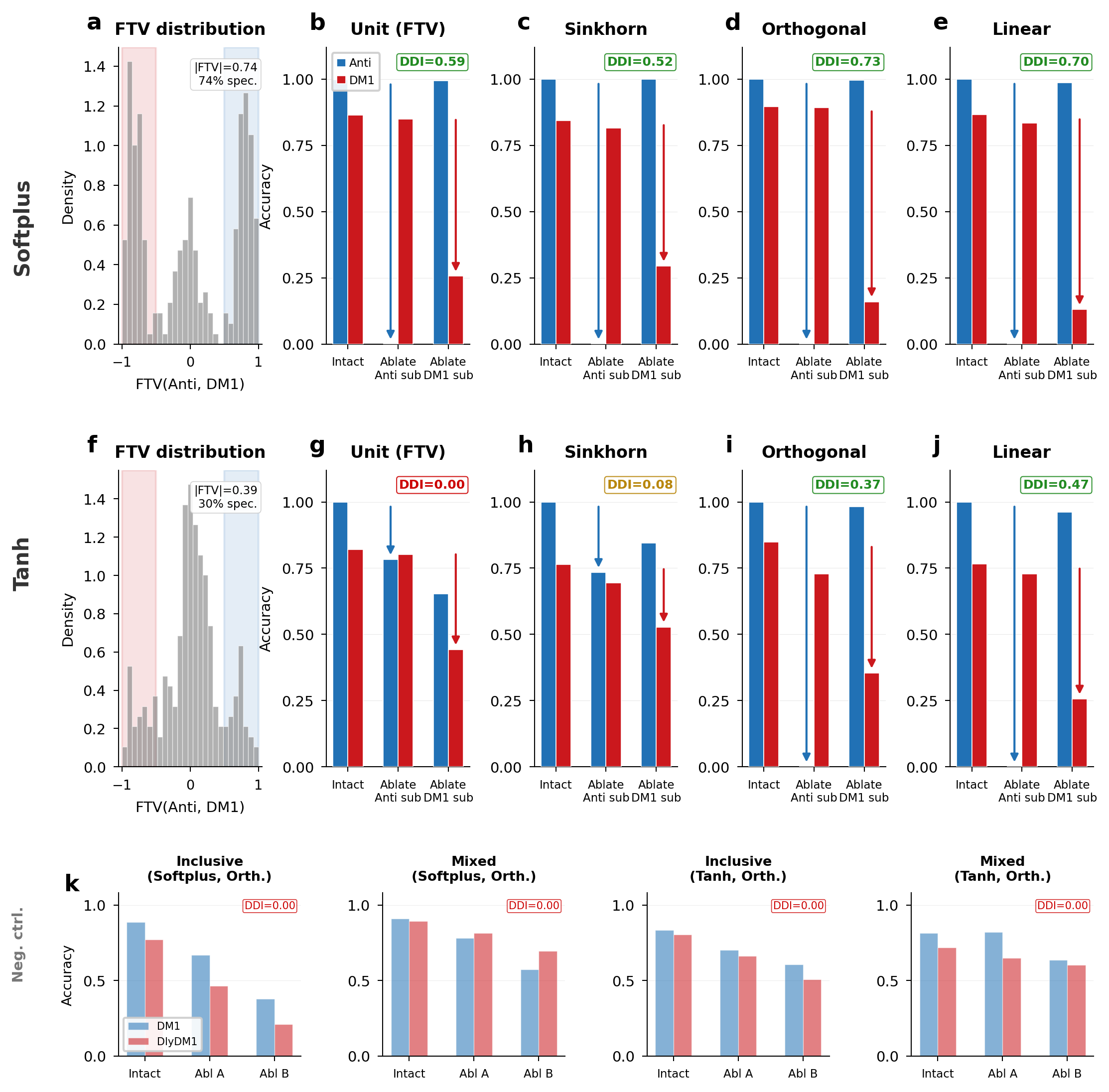}
\caption{\textbf{Functional modularity in multitask RNNs (Yang et al., 2019).}
\textbf{(a,f)}~FTV distributions for Anti vs DM1 in softplus (top) and tanh (bottom) networks. Softplus shows bimodal clustering; tanh shows a broad unimodal distribution centered near zero.
\textbf{(b--e)}~Double dissociation bars for softplus networks at each hierarchy level (unit, Sinkhorn, orthogonal, linear). All levels show selective ablation effects (DDI = 0.59--0.73).
\textbf{(g--j)}~Same for tanh networks. Unit-level DDI=0.00; orthogonal DDI = 0.37; linear DDI = 0.47. Subspace decomposition recovers modularity that is invisible at the unit level.}
\label{fig:yang}
\end{figure}

We applied our hierarchy to networks trained on 11 cognitive tasks spanning sensorimotor mapping, perceptual decision-making, context-dependent integration, working memory, and categorical matching \citep{yang2019task}. We trained networks with both softplus and tanh activations. Following the taxonomy of \citet{yang2019task}, we evaluate Anti vs DM1 (disjoint) task pair that span the spectrum from fully separable to fully shared computation. Anti requires responding in the opposite direction of a single stimulus (sensorimotor remapping), while DM1 requires reporting which of two overlapping stimuli is stronger (perceptual evidence comparison). These tasks share no computational requirements.

Figure~\ref{fig:yang} first shows results for the disjoint pair (Anti vs DM1). We begin with the fractional task variance (FTV) analysis introduced by \citet{yang2019task}, which measures, for each unit, the relative variance contributed by each task. In softplus networks, the FTV distribution is strongly bimodal ($|\text{FTV}| = 0.74$, with 74\% of units specialized for one task), indicating that Anti and DM1 are supported by largely non-overlapping populations (Figure~\ref{fig:yang}a). In tanh networks, the FTV distribution is broad and unimodal ($|\text{FTV}| = 0.39$, 30\% specialized), with most units contributing similarly to both tasks (Figure~\ref{fig:yang}f). This is the observation that led \citet{yang2019task} to conclude that tanh networks lack modular structure and that ``state-space approaches may be more appropriate'' for their analysis.

Our decomposition hierarchy reveals interesting trends. In softplus networks, all levels recover a double dissociation, with DDI increasing from unit (0.59) through orthogonal (0.73) to linear (0.70) (Figure~\ref{fig:yang}b--e). The strong FTV clustering means that even the simplest decomposition (assigning units to tasks based on their FTV values) produces selective ablation effects, consistent with the cluster-based lesioning results reported by \citet{yang2019task}. The critical result is in tanh networks. Unit-level decomposition yields DDI $= 0.00$: despite the FTV identifying some weakly specialized units, ablating them produces collateral damage that overwhelms their selective effects (Figure~\ref{fig:yang}g). Sinkhorn permutation, which searches for a better unit assignment than FTV ranking, yields only DDI $= 0.08$. But orthogonal decomposition achieves DDI $= 0.37$ (Figure~\ref{fig:yang}i), and linear decomposition reaches DDI $= 0.47$ (Figure~\ref{fig:yang}j), recovering a clear double dissociation in a network where unit-level analysis finds no modular structure. This confirms the hypothesis of \citet{driscoll2024flexible} that such networks possess functional structure ``not aligned to unit axes.''

More experiments on Multitask cognitive RNNs \citet{mante2013context} are in Appendix section \ref{subsec:Mante task}. Together, the Mante and Yang results establish two key findings. First, the modularity hierarchy reveals structure that unit-level analyses miss: networks that appear non-modular under conventional analysis harbor clear functional modularity at the subspace level. Second, the level at which modularity emerges depends on the network's computational organization: softplus networks are modular at all levels including unit, while tanh networks require orthogonal or linear decomposition to expose the same functional separation.

\subsection{Content-Style Modularity in Vision Models}
\label{sec:vision_results}
We apply our modularity hierarchy to the question of whether content and style occupy functionally separable subspaces in pretrained vision models. Content refers to structural and semantic properties of the image (object identity, shape, 
spatial layout) while style refers to surface-level properties such as texture, color palette, and artistic rendering. We train decompositions using the contrastive minimal-pairs objective defined above and evaluate the resulting subspaces through retrieval-based 
selectivity metrics on held-out data. Our test set is constructed so that every image has exactly 99 other images sharing the same content (different stylizations 
of the same scene) and 99 sharing the same style (different scenes rendered in the same artistic style) (Appendix section \ref{subsec:datasets}). This structure provides ground truth for evaluating subspace quality without additional annotation. We assess modularity through two 
complementary analyses.

\textbf{Correlation analysis.} We partition test images into style-grouped sets $\mathcal{G}^s$ (100 images sharing the same style, varying in content) and content-grouped sets $\mathcal{G}^c$ (100 images sharing the same content, varying in style). For each grouping, we compute the average pairwise Pearson correlation of representations within each subspace. A modular decomposition should produce $\rho(\mathbf{z}_{\text{style}} \mid \mathcal{G}^s) \gg \rho(\mathbf{z}_{\text{style}} \mid \mathcal{G}^c)$ and $\rho(\mathbf{z}_{\text{content}} \mid \mathcal{G}^c) \gg \rho(\mathbf{z}_{\text{content}} \mid \mathcal{G}^s)$ simultaneously.

\textbf{Retrieval selectivity.} For each query image, we retrieve its top-$k$ nearest neighbors in each subspace and measure the fraction sharing the correct property. We define the content metric $\mathrm{CM} = (C_c - S_c) / C_c$ and style metric $\mathrm{SM} = (S_s - C_s) / S_s$, where $C_c$ is content retrieval accuracy in the content subspace and $S_c$ in the style subspace, and analogously for style. We additionally evaluate selectivity at the level of individual 
images. For each test image $x_i$, let $\mathcal{N}^s_{99}(x_i)$ and $\mathcal{N}^c_{99}(x_i)$ denote its 99 nearest neighbors in the style and content subspaces respectively (under Euclidean distance), and let $\mathcal{S}(x_i)$ and $\mathcal{C}(x_i)$ denote all images sharing its style or content. We define four 
per-image selectivity scores: $P_{ss}(x_i) = \frac{|\mathcal{N}^s_{99}(x_i) \cap \mathcal{S}(x_i)|}{99}$, $P_{cs}(x_i) = \frac{|\mathcal{N}^c_{99}(x_i) \cap \mathcal{S}(x_i)|}{99}$,  $P_{cc}(x_i) = \frac{|\mathcal{N}^c_{99}(x_i) \cap \mathcal{C}(x_i)|}{99}$ and $P_{sc}(x_i) = \frac{|\mathcal{N}^s_{99}(x_i) \cap \mathcal{C}(x_i)|}{99}$, 
where the first subscript denotes the subspace used for retrieval and the second denotes the property being evaluated. $P_{ss}$ measures how many of an image's style-subspace neighbors share 
its style (desired high), while $P_{sc}$ measures contamination of the style subspace by content (desired low), and symmetrically 
for $P_{cc}$ and $P_{cs}$. We summarize these as selectivity 
ratios: $R_s = \frac{\mathbb{E}[P_{ss}]}{\mathbb{E}[P_{cs}]}$ and $R_c = \frac{\mathbb{E}[P_{cc}]}{\mathbb{E}[P_{sc}]}$, where $R_s, R_c \gg 1$ jointly indicate that both subspaces are selective for their intended property.


\paragraph{Results}\label{subsec:vision_results_sanity_check}





\begin{figure}[h!]
\centering
\begin{minipage}{0.48\linewidth}
    \centering
    \includegraphics[width=\linewidth]{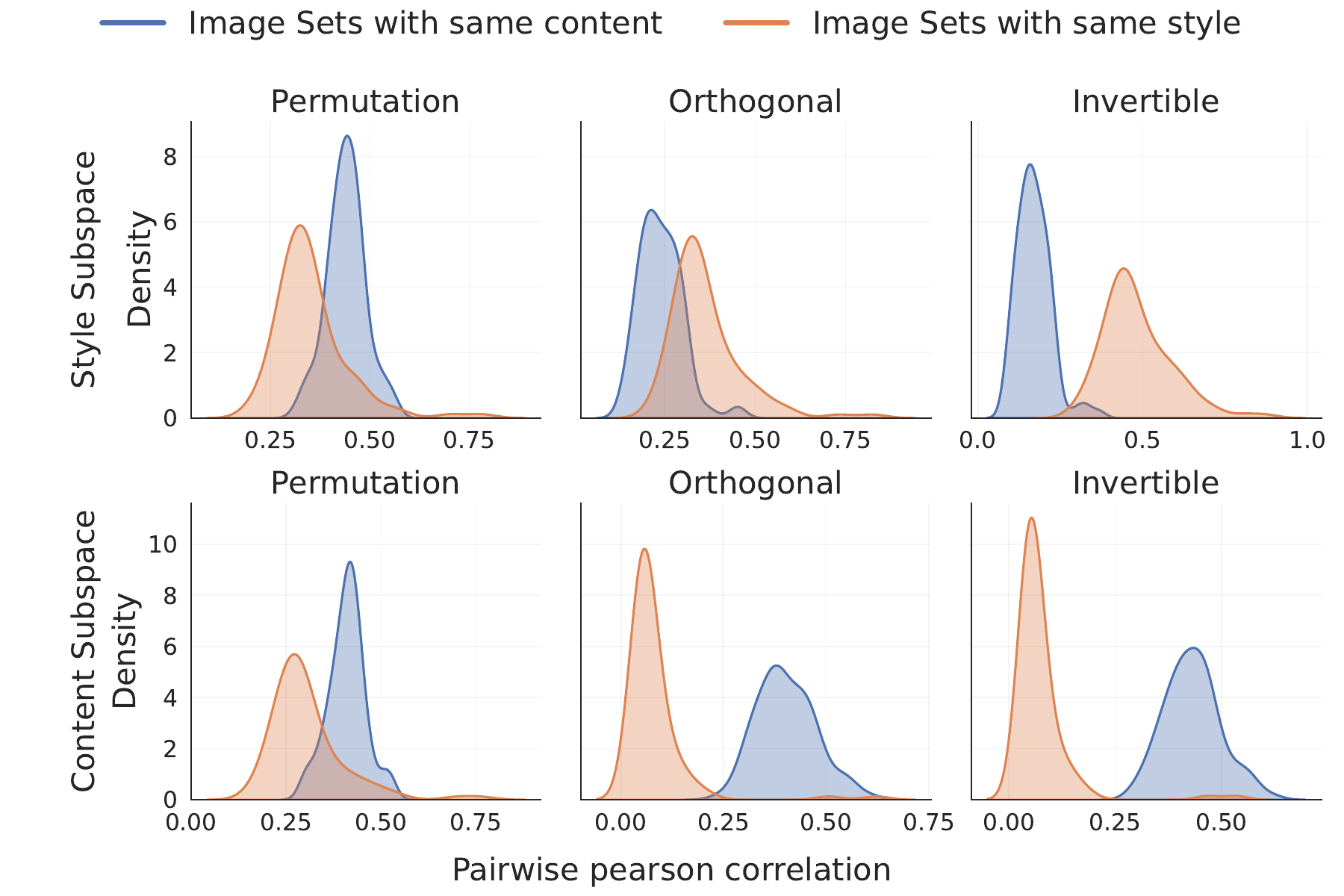}
    \caption{Swin Base: For each subspace (style/content), we plot the distribution of average pairwise Pearson correlations computed within sets of images grouped by shared style or content, across permutation, orthogonal, and invertible transforms ($\rho(\mathbf{z}_{\text{style}}\mathcal{G}^s)$, $\rho(\mathbf{z}_{\text{style}} \mid \mathcal{G}^c)$, $\rho(\mathbf{z}_{\text{content}} \mid \mathcal{G}^c)$ and $\rho(\mathbf{z}_{\text{content}} \mid \mathcal{G}^s)$).}
    \label{fig:distribution_comp_main}
\end{minipage}
\hfill
\begin{minipage}{0.48\linewidth}
    \centering
    \includegraphics[width=\linewidth]{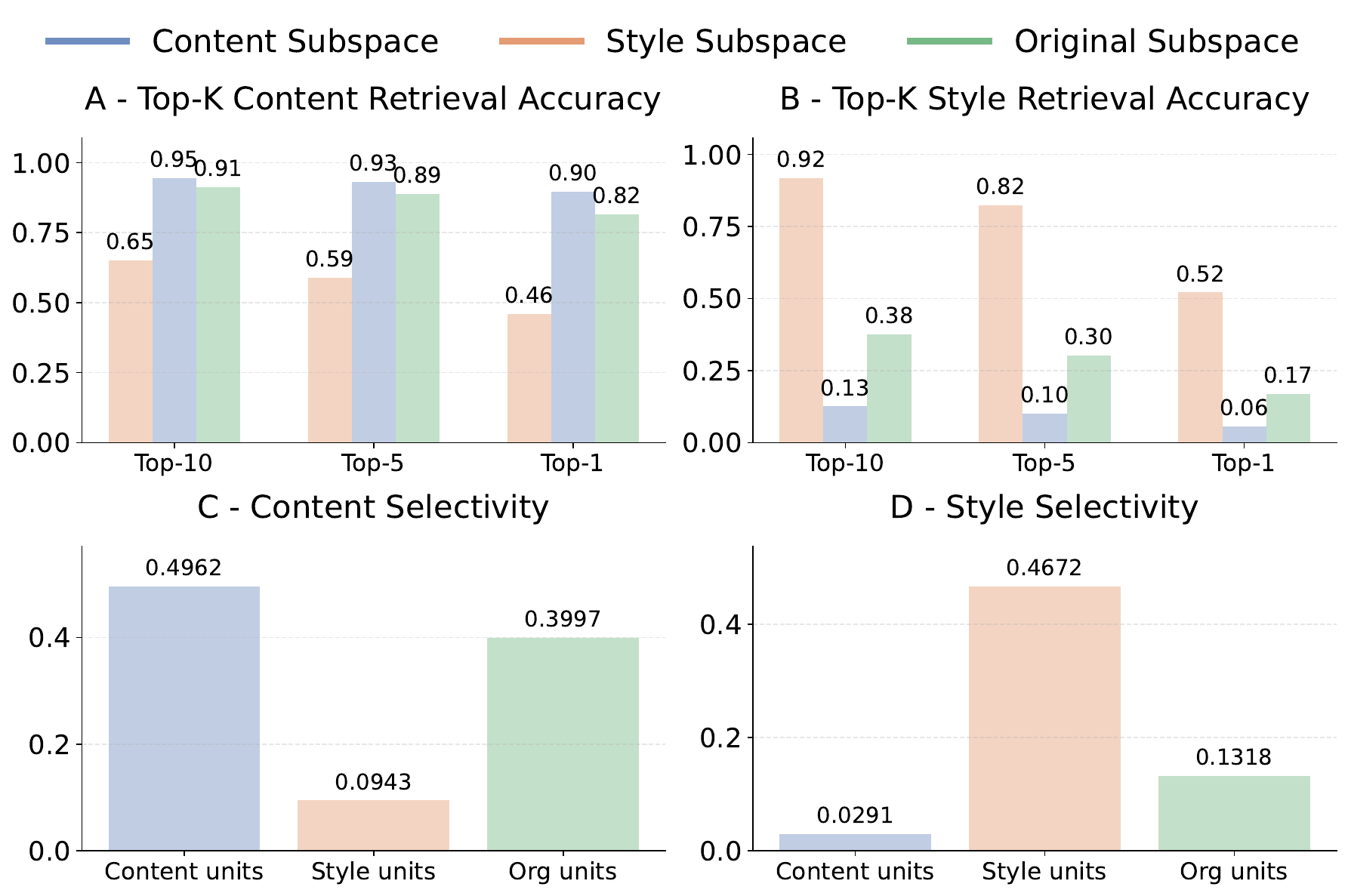}
    \caption{Swin Base: (A) Top-K content retrieval ($C_c$, $S_c$); (B) Top-K style retrieval ($C_c$, $S_c$); (C) Content selectivity ($P_{cc}$, $P_{sc}$); and (D) Style selectivity $P_{ss}$, $P_{cs}$ —each evaluated across content and style subspaces under invertible transforms.}
    \label{fig:metric_plots_swinb}
\end{minipage}

\end{figure}

Figure~\ref{fig:distribution_comp_main} shows the correlation analysis for the Swin-Base model. In all cases, style-grouped images yield higher correlations in the style subspace than in the content subspace, and content-grouped images yield higher correlations in the content subspace, confirming that the learned partition captures the intended functional distinction. The separation increases progressively from permutation through orthogonal to invertible transformations. Detailed comparisons on other model architectures can be found in Appendix figures \ref{fig:distribution_invertible}, \ref{fig:distribution_orthogonal} and \ref{fig:distribution_permutation}. Further Appendix figure \ref{fig:umap} shows that images sharing the same functional attribute form coherent, well-defined clusters within the corresponding functional subspace manifold, indicating that the decomposition induces meaningful geometric organization aligned with the target attributes.

\begin{table}[ht]
\centering
\resizebox{\columnwidth}{!}{%
\begin{tabular}{@{}lrrrrrrrr@{}}
\toprule
\textbf{Transformation}
    & \textbf{Top-1 CM} $\uparrow$
    & \textbf{Top-5 CM} $\uparrow$
    & \textbf{Top-10 CM} $\uparrow$
    & \textbf{Top-1 SM} $\uparrow$
    & \textbf{Top-5 SM} $\uparrow$
    & \textbf{Top-10 SM} $\uparrow$
    & \textbf{$R_s$ $\uparrow$}
    & \textbf{$R_c$ $\uparrow$} \\
\midrule
Permutation & 0.0063 & 0.0080 & 0.0125 & 0.0738 & 0.0625 & 0.0546 & 1.1032 & 1.0216 \\
Orthogonal  & 0.1381 & 0.1686 & 0.2435 & 0.7542 & 0.7550 & 0.7613 & 7.2348 & 2.5603 \\
Invertible  & \textbf{0.3107} & \textbf{0.3675} & \textbf{0.4884} & \textbf{0.8632} & \textbf{0.8785} & \textbf{0.8940} & \textbf{16.0815} & \textbf{5.2594} \\
\bottomrule
\end{tabular}%
}
\caption{Swin Base - Comparison of Top-K content and style summary metrics, along with content and style selectivity ratios, across permutation, orthogonal, and invertible transformations.}
\label{tab:table_swinb_baseline_comp}
\end{table}

\begin{figure}[h!]
\centering
\includegraphics[width=0.9\linewidth]{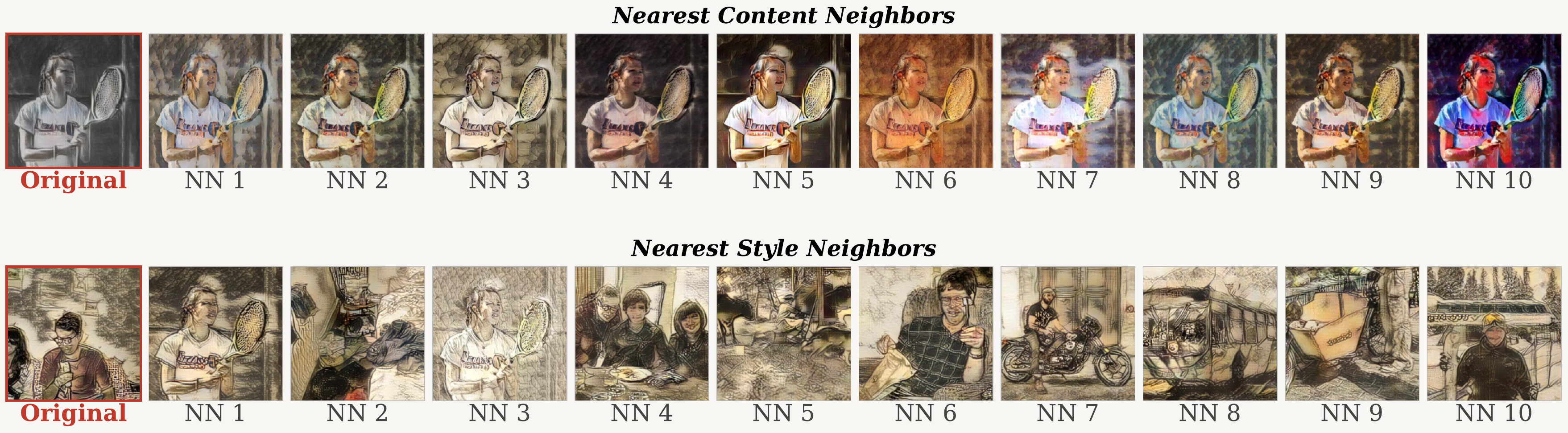}
\caption{Top-10 nearest neighbor retrieval results under invertible transforms in the content subspace and style subspace}
\label{fig:retrieval_grid}
\end{figure}

Table~\ref{tab:table_swinb_baseline_comp} reports the retrieval metrics. The same hierarchy holds: permutation-based decompositions achieve near-chance selectivity ($R_s \approx 1$, $R_c \approx 1$), indicating that content and style are not separable at the unit level. Orthogonal decompositions yield moderate selectivity, and invertible transformations achieve strong separation ($R_s = 16.1$, $R_c = 5.3$). This progression mirrors the modularity hierarchy: content and style are not axis-aligned (permutation fails), not orthogonally separable (orthogonal is moderate), but are separable under a general linear transformation. 
Figure~\ref{fig:metric_plots_swinb}A--B shows retrieval 
accuracies for the Swin-Base transformer under the invertible 
transform, broken down by subspace and property: the content subspace retrieves content matches more accurately than the style subspace ($C_c > S_c$), and vice versa for style 
($S_s > C_s$), confirming that each subspace is selective for 
its intended property. Selectivity scores (Figure~\ref{fig:metric_plots_swinb}C--D) corroborate this 
pattern, and Figure~\ref{fig:retrieval_grid} provides a qualitative illustration, showing nearest neighbors retrieved from each subspace for an example query. 

This pattern generalizes across architectures. Across Swin transformers, ConvNeXts, VGGs, and self-supervised ViTs, invertible transforms consistently achieve the strongest content--style separation (Appendix Figures~\ref{fig:distribution_invertible}--\ref{fig:distribution_permutation}). The principal exception is the ResNet family, where style retrieval is weak across all subspace types, suggesting that these models encode style information in a form less amenable to linear separation. Additional evaluation on conflicting-cue stimuli and silhouette--edge stimuli from \citet{geirhos2018imagenet} further confirms that the content subspace is selective for shape-based structure while the style subspace captures texture-based properties (Appendix~\ref{subsec:geirhos_silhoutte}, \ref{subsec:geirhos_conflicting_cues}). Appendix Section~\ref{subsec:ablation_transformation_flexibility} provides additional ablation and control experiments demonstrating that flexible learned transformations reveal modular structure already present in the representation, rather than manufacturing task-specific modularity through arbitrary reparameterization.

\subsection{Syntax--Semantics Modularity in Language Models}
\label{sec:language_results}
We apply the same modularity hierarchy to language model 
representations, asking whether syntactic and semantic 
processing occupy separable subspaces. Decompositions are 
trained using the contrastive minimal-pairs objective (Eq.~\ref{eq:loss_contrastive}) on sentences from the 
Microsoft Research Paraphrase Corpus and evaluated on the 
Pereira dataset (see Appendix~\ref{subsec:datasets} for details).

\begin{table}[ht]
\centering
\resizebox{\columnwidth}{!}{%
\begin{tabular}{@{}lrrrrrrrr@{}}
\toprule
\textbf{Transformation}
    & \textbf{Top-1 $S_1$} $\uparrow$
    & \textbf{Top-5 $S_1$} $\uparrow$
    & \textbf{Top-10  $S_1$} $\uparrow$
    & \textbf{Top-1  $S_2$} $\uparrow$
    & \textbf{Top-5  $S_2$} $\uparrow$
    & \textbf{Top-10 $S_2$} $\uparrow$ \\
\midrule
Permutation & 0.3333  & 0.3953  & 0.3971  & 0.1564  & 0.1681  & 0.1368 \\
Orthogonal  & 0.8485  & 0.9286  & 0.9417  & 0.6667  & 0.5504  & 0.4763 \\
Invertible  & \textbf{0.9375}  & \textbf{0.9439}  & \textbf{0.9259}  & \textbf{0.6556}  & \textbf{0.5675}  & \textbf{0.5152} \\
\bottomrule
\end{tabular}%
}
\caption{Comparison of Top-K Syntax and Semantics summary metrics across permutation, orthogonal, and invertible transformations for Pythia 410M.}
\label{tab:table_gpt2xl_baseline_comp}
\end{table}

\textbf{Evaluation - }For each sentence, we construct a semantic retrieval task with one 
semantic counterpart (a paraphrase preserving meaning while 
altering surface form) and 70 syntactic counterparts (sentences 
sharing the same grammatical template with different lexical 
content) and a syntactic retrieval task with one syntactic counterpart and 70 semantic counterparts. We perform top-$k$ nearest-neighbor retrieval in both 
the syntactic and semantic subspaces and measure whether the 
correct counterpart type is retrieved. We define syntax and 
semantics selectivity metrics analogous to the vision case: $S_1 = \frac{\mathrm{Syn}_{\mathrm{Syn}} - \mathrm{Sem}_{\mathrm{Syn}}}{\mathrm{Syn}_{\mathrm{Syn}}}$ and $S_2 = \frac{\mathrm{Sem}_{\mathrm{Sem}} - \mathrm{Syn}_{\mathrm{Sem}}}{\mathrm{Sem}_{\mathrm{Sem}}}$,
where $\mathrm{Syn}_{\mathrm{Syn}}$ is the retrieval accuracy 
of the syntactic subspace on syntactic counterparts, 
$\mathrm{Sem}_{\mathrm{Syn}}$ is the retrieval accuracy of the 
semantic subspace on syntactic counterparts, and symmetrically 
for $\mathrm{Sem}_{\mathrm{Sem}}$ and $\mathrm{Syn}_{\mathrm{Sem}}$. $S_1$ measures the advantage of the syntactic 
subspace over the semantic subspace at retrieving structurally 
matched sentences, and $S_2$ measures the analogous advantage 
for meaning-matched retrieval.

\textbf{Results - } Table~\ref{tab:table_gpt2xl_baseline_comp} presents these results for GPT-2 XL. Results across a broader range of model sizes (from 160M to 1.5B parameters) are provided in Appendix Table~\ref{tab:language_baseline_comp}, including Pythia 160M and 410M models. Consistent with the vision findings, invertible linear transformations yield the highest degree of subspace modularity, followed by orthogonal transformations, while permutation-based transformations perform the worst.

Finally, we examine semantic and syntactic selectivity for Pythia 410M under the invertible transformation (Figure~\ref{fig:syn_sem_pythia_410m}; results for other models are shown in Appendix Figures~\ref{fig:syn_sem_pythia_160m} and~\ref{fig:syn_sem_gpt2xl}). We observe strong evidence of disentanglement between syntax and semantics. The strong semantic disentanglement is not entirely surprising, as paraphrases in the dataset often share many common tokens with their semantic counterparts. More strikingly, the syntactic subspaces capture syntactic structure with a high degree of fidelity, despite the substantial variability in surface form and the minimal overlap in tokens across examples that share the same underlying structure (Appendix Figure \ref{fig:token_overlap}). Both semantic and syntactic selectivity metrics remain robust across all different layers of a model(Appendix Figure~\ref{fig:all_layers_comp}), indicating that disentanglement persists throughout the full depth of the network rather than being confined to specific layers.

\begin{figure}[h!]
\centering

\begin{minipage}{0.49\linewidth}
    \centering
    \includegraphics[width=\linewidth]{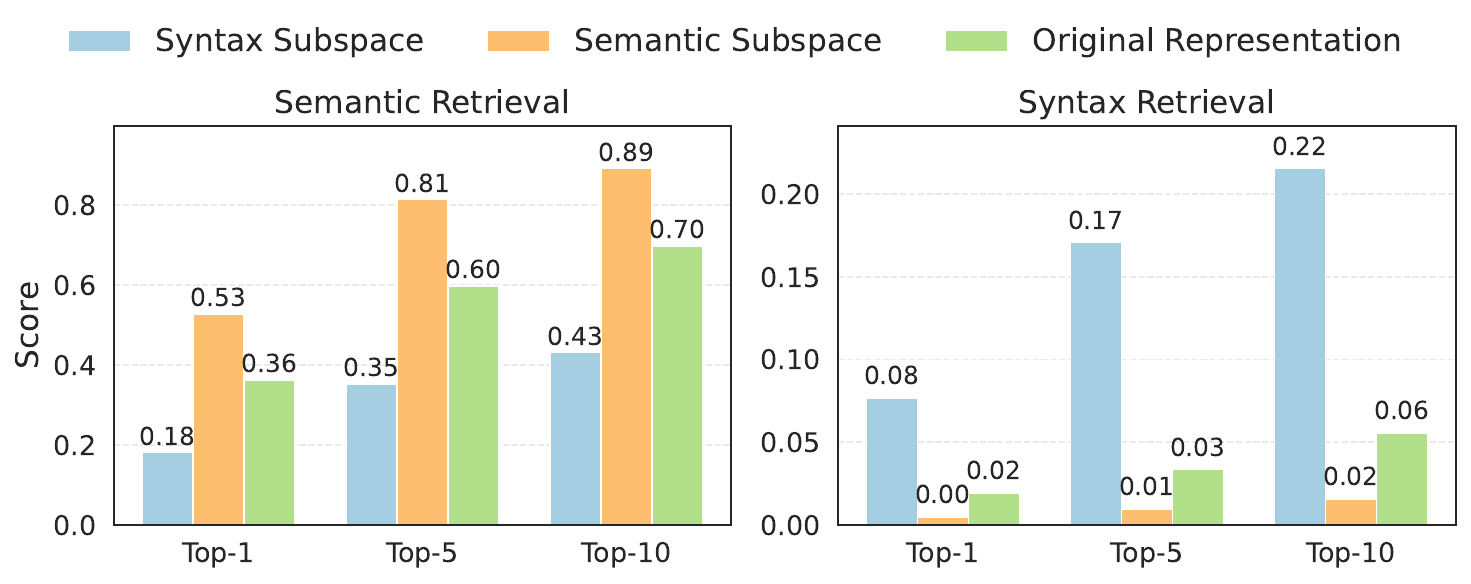}
    \captionof{figure}{Top-K Semantic ($\mathrm{Syn}_{\mathrm{Sem}}$, $\mathrm{Sem}_{\mathrm{Sem}}$) \& Syntax retrieval ($\mathrm{Syn}_{\mathrm{Syn}}$, $\mathrm{Sem}_{\mathrm{Syn}}$) for Pythia 410M}
    \label{fig:syn_sem_pythia_410m}
\end{minipage}
\hfill
\begin{minipage}{0.49\linewidth}
    \centering
    \includegraphics[
        width=\linewidth,
        trim=0 6cm 3cm 0,
        clip
    ]{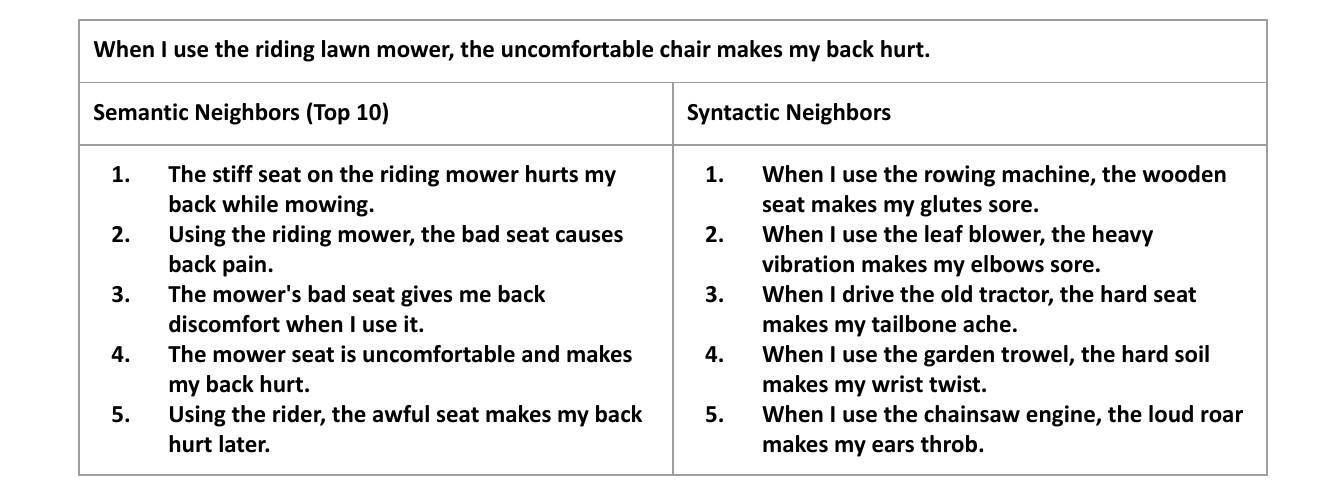}
    \captionof{figure}{Top-5 nearest neighbor retrieval in syntax and semantic subspaces for Pythia 410M}
    \label{fig:language_nn}
\end{minipage}

\label{fig:ss}

\end{figure}

\begin{figure}[t]
\centering
\includegraphics[width=0.88\textwidth]{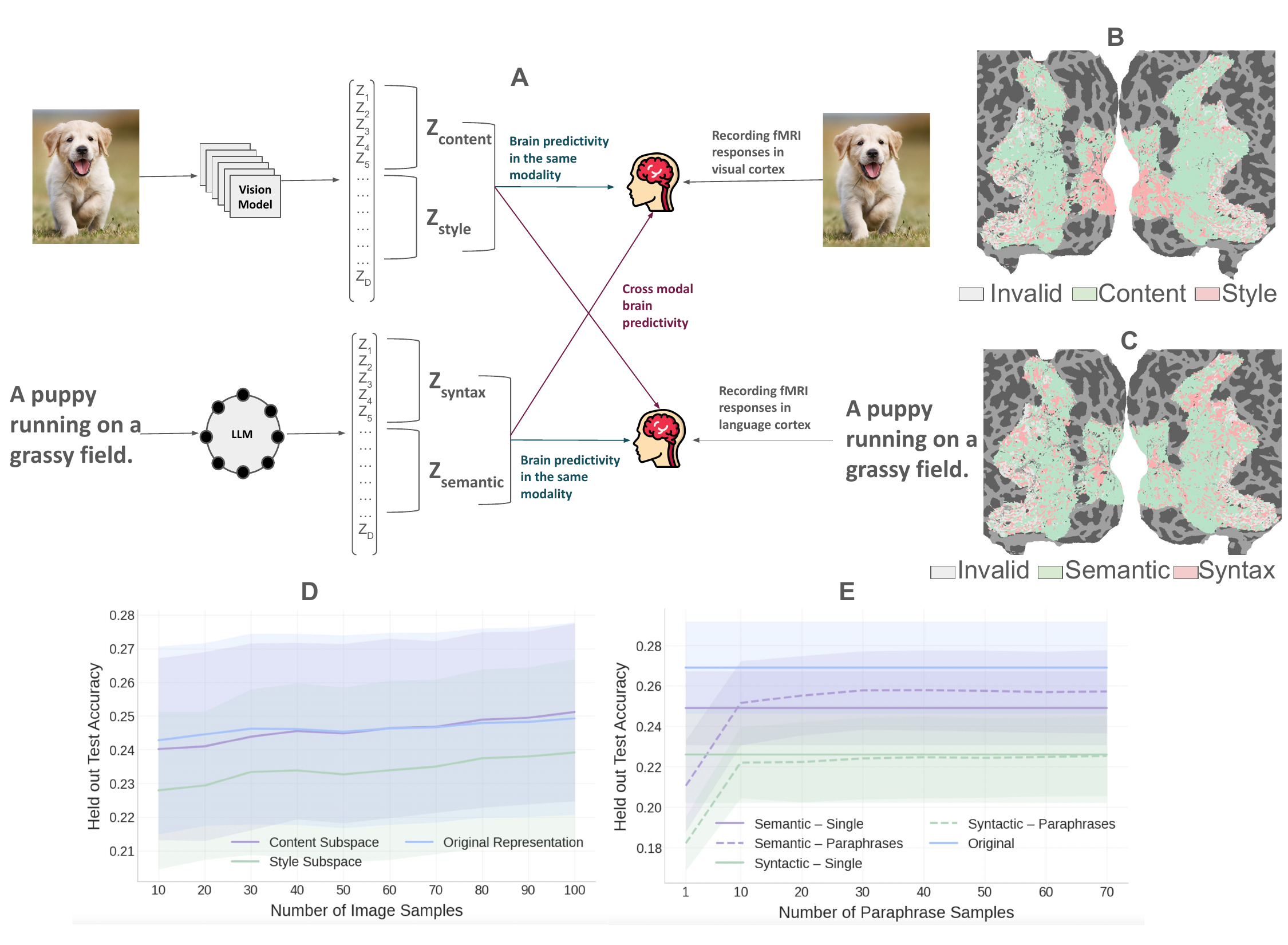}
D\caption{\textbf{(A)} Schematic illustrating how distinct subspace representations derived from vision and language models are used to predict neural activity in the human visual and language cortex.
\textbf{(B)} Regions of the visual cortex that are better predicted by style versus content subspace representations from vision models.
\textbf{(C)} Regions of the visual cortex that are better predicted by syntactic versus semantic subspace representations from language models.
\textbf{(D)} Predictive performance of style and content subspace representations from vision models in explaining neural activity within the language cortex.
\textbf{(E)} Predictive performance of syntactic and semantic subspace representations from language models in explaining neural activity within the language cortex.}
\label{fig:combined_brain_subspace}
\end{figure}

\subsection{Neural Alignment of Learned Subspaces}
\label{subsec:brain}

A central question in computational neuroscience is why certain artificial neural network representations predict brain activity better than others \citep{yamins2014performance, schrimpf2020integrative}. If a representation can be decomposed into functionally distinct subspaces, we can ask a more refined question: which component of the representation drives the alignment? We apply our learned decompositions to neural encoding models to test this.





\textbf{Content and style in visual cortex.}
We train encoding models to predict fMRI responses in the 
Natural Scenes Dataset (\citealt{allen2022massive}), using content and style subspace representations as separate feature sets (Figure \ref{fig:combined_brain_subspace}-A). Encoding models are fit and evaluated on the 1{,}000 
images viewed by all four subjects (1, 2, 5, and 7). Early visual areas (V1, V2) are better predicted by the style 
subspace, while higher visual regions associated with object 
recognition are better predicted by the content subspace 
(Figure~\ref{fig:combined_brain_subspace}B). This dissociation 
mirrors the established cortical hierarchy: early areas process 
low-level features such as texture and local statistics, while later areas encode semantic information~\citep{saha2024modeling, yamins2014performance, tang2023brain, wang2022incorporating, doerig2022visual}. 

\textbf{Syntax and semantics in language cortex.}
We next evaluate syntactic and semantic language representations as predictors of activity in the language cortex (Figure \ref{fig:combined_brain_subspace}-A), defined functionally via the words-versus-nonwords localizer contrast \citep{fedorenko2010new} (Figure Figure \ref{fig:combined_brain_subspace}-B). The semantic subspace outperforms the syntactic subspace, consistent with evidence that the language network may be generally more strongly concerned with meaning than syntactic form~\citep{fedorenko2020lack} (Figure \ref{fig:combined_brain_subspace}-E). Notably, the original sentence representations which integrate both syntactic and semantic information yield the strongest overall brain predictivity, suggesting that neither dimension alone fully captures the representations encoded in the language network. 

\textbf{Cross-modal model-brain comparisons}
Finally, we demonstrate that the semantic advantage generalizes across modalities. Semantic representations derived from image captions predict visual cortex activity better than their syntactic counterparts, further underscoring the privileged role of meaning-level structure in driving brain alignment (Figure~\ref{fig:combined_brain_subspace}-A,C). Next, we generate images from sentence stimuli (Pereira dataset) and extract content and style subspace representations, averaging across different number of images per stimulus. Content representations consistently outperform style representations in predicting language cortex activity (Figure~\ref{fig:combined_brain_subspace}-A,D), and their performance closely matches that of the original sentence representations suggesting that semantic content, rather than low-level stylistic features, accounts for the bulk of image-to-language-cortex alignment.

\section{Discussion}
We introduced a hierarchy of three nested notions of modularity and showed that networks widely considered non-modular often 
harbor clear functional modularity at the subspace level. Different levels of modularity likely carry distinct functional 
consequences. Unit-level modularity permits low wiring costs:  downstream circuits that depend on a specific function need only connect to the dedicated neurons. It also underwrites the classical neuropsychological inference: spatially localized damage spares unrelated functions because co-located neurons serve the same function. In artificial networks, where neurons lack spatial organization, 
the practical significance shifts: subspace-level modularity 
means that specific computations can be selectively accessed and 
intervened upon through a known linear transformation, even when 
no individual neuron is dedicated to any single function. Characterizing further implications for robustness, 
generalization, and learning dynamics across all three 
levels remains an open question.

More broadly, decomposing representations into functional subspaces opens new 
analytical possibilities: our neural alignment results 
illustrate that one can ask which functional component of a 
representation drives alignment with brain data, rather than 
treating the representation as monolithic. This may also refine questions about cross-model convergence 
\citep{huh2024platonic}: if models trained on different 
objectives and modalities converge in their representations, 
subspace decomposition can reveal whether this convergence is 
driven by specific functional components, such as shared 
semantic structure, rather than the representation as a 
whole. Several further 
questions remain unexplored, including how subspace structure 
and dimensionality evolve across training and across layers, and whether the 
hierarchy extends naturally to multi-way decompositions beyond 
two functions. An additional open question concerns the role of the residual subspace. As shown in Appendix Section \ref{subsec:app_residual_subspace}, the residual is not purely nuisance: it can contain or support task-relevant information, but this information appears to be shared or redundant rather than selectively associated with either task. Understanding the functional organization and role of this residual information remains an open direction for future work.

Our approach has practical limitations. Learning decompositions requires non-convex optimization with sensitivity to hyperparameters (subspace dimensionality, condition number bound, learning rate). For the vision and language experiments, we evaluate subspace quality through retrieval-based selectivity. A natural next step is to complement these with behavioral evaluations, for instance, measuring whether projecting out the content subspace selectively impairs object recognition while 
preserving texture judgments, as we demonstrated for the 
RNN experiments. A further open question concerns the role of the residual units: it remains unclear what information they retain and how it is organized, which we leave for future work. More broadly, 
we believe the question of at what level of description 
modularity exists, and what each level implies, will be 
a productive line of inquiry for both neuroscience and 
interpretability research.

\bibliographystyle{plainnat}
\bibliography{example_paper}

\newpage
\renewcommand{\thesection}{\Alph{section}}
\renewcommand{\thesubsection}{\Alph{section}.\arabic{subsection}}

\renewcommand{\thefigure}{\Alph{section}\arabic{figure}}
\renewcommand{\thetable}{\Alph{section}\arabic{table}}

\setcounter{figure}{0}
\setcounter{table}{0}
\section{Appendix}

\subsection{Differentiable Constraints via GeoTorch and Sinkhorn Relaxation}
\label{subsec:constraints}

We imposes geometric constraints on neural network parameters through a
\textit{parametrization} strategy (use the Geotorch library in the case of orthogonal and invertible transformations): rather than optimizing constrained parameters
directly, we maintains unconstrained parameters internally and maps them through
a differentiable, manifold-specific function every time the weight is accessed.

\subsubsection{Orthogonal Constraint (\texttt{geotorch.orthogonal})}

The target manifold is the Special Orthogonal group
\[
  \mathrm{SO}(n) \;=\; \bigl\{\, Q \in \mathbb{R}^{n \times n}
    \;\mid\; Q^{\top}Q = I,\; \det(Q) = 1 \,\bigr\},
\]

\paragraph{Mechanism.}
Let $A \in \mathbb{R}^{n \times n}$ be the unconstrained raw parameter stored by
the optimizer. GeoTorch first forms a skew-symmetric matrix
\[
  B \;=\; A - A^{\top}, \qquad B^{\top} = -B,
\]
and then applies a \emph{trivialization} map $\phi: \mathfrak{so}(n) \to
\mathrm{SO}(n)$ to produce a valid orthogonal matrix. The default trivialization
is the \textbf{matrix exponential}:
\[
  Q \;=\; \exp(B).
\]
This is the Lie-group exponential map from the Lie algebra $\mathfrak{so}(n)$
(skew-symmetric matrices) to the Lie group $\mathrm{SO}(n)$; it is surjective,
so every orthogonal matrix is reachable. An alternative trivialization is the
\textbf{Cayley map}
\[
  Q \;=\; (I - B)(I + B)^{-1},
\]
which requires only a linear solve and is cheaper for large $n$, though it
cannot represent orthogonal matrices with eigenvalue $-1$.

\subsubsection{Linear Invertible Constraint (\texttt{geotorch.invertible})}

The target manifold is the General Linear group with positive determinant,
\[
  \mathrm{GL}^{+}(n) \;=\; \bigl\{\, W \in \mathbb{R}^{n \times n}
    \;\mid\; \det(W) > 0 \,\bigr\}.
\]

\paragraph{Mechanism.}
The constraint is imposed via an SVD-based parametrization with three
components:

\begin{enumerate}
  \item \textbf{Orthogonal factors} $U, V \in \mathrm{SO}(n)$: two independent
    skew-symmetric raw parameters $A_U, A_V$ are each passed through the matrix
    exponential (default) or Cayley map, exactly as in the orthogonal case.

  \item \textbf{Singular values} $\sigma_i > 0$: a raw vector
    $s \in \mathbb{R}^n$ is passed through the \texttt{softplus} function,
    \[
      \sigma_i \;=\; \log\!\bigl(1 + e^{s_i}\bigr) \;>\; 0
      \quad \forall\, s_i \in \mathbb{R},
    \]
    which maps all of $\mathbb{R}$ into $(0, \infty)$, guaranteeing strict
    positivity.
\end{enumerate}

The constrained weight is then reconstructed as
\[
  W \;=\; U\,\mathrm{diag}(\boldsymbol{\sigma})\,V^{\top},
\]
which satisfies $\det(W) = \det(U)\det(V)\prod_i \sigma_i > 0$ by construction.

\subsubsection{Permutation Constraint (Sinkhorn Relaxation)}

Direct optimization over the set of $d \times d$ permutation matrices
\[
  \mathcal{P}_d \;=\; \bigl\{\, P \in \{0,1\}^{d \times d}
    \;\mid\; P\mathbf{1} = \mathbf{1},\; P^{\top}\mathbf{1} = \mathbf{1} \,\bigr\}
\]
is combinatorial and therefore not amenable to gradient-based optimization.
We instead adopt a differentiable relaxation over the Birkhoff polytope of
doubly-stochastic matrices,
\[
  \mathcal{B}_d \;=\; \bigl\{\, S \in \mathbb{R}_{\geq 0}^{d \times d}
    \;\mid\; S\mathbf{1} = \mathbf{1},\; S^{\top}\mathbf{1} = \mathbf{1} \,\bigr\},
\].

\paragraph{Mechanism.}
Let $A \in \mathbb{R}^{d \times d}$ be an unconstrained parameter matrix. We
define the relaxed permutation as
\[
  S \;=\; \mathrm{Sinkhorn}\!\left(\exp\!\left(A\right)\right),
\]
where $\mathrm{Sinkhorn}(\cdot)$
denotes iterative row and column normalization, which converges to a
doubly-stochastic matrix. Concretely, starting from
$M^{(0)} = \exp(A )$, the iterations are
\[
  M^{(t+\frac{1}{2})}_{ij} \;=\; \frac{M^{(t)}_{ij}}{\sum_{j'} M^{(t)}_{ij'}},
  \qquad
  M^{(t+1)}_{ij} \;=\; \frac{M^{(t+\frac{1}{2})}_{ij}}{\sum_{i'} M^{(t+\frac{1}{2})}_{i'j}},
\]
alternating row and column normalization until convergence. The resulting $S$
satisfies all row and column sum constraints by construction and is fully
differentiable with respect to $A$, admitting gradient flow through the
normalization steps.

\paragraph{Inference.}
After training, $S$ is projected onto the nearest permutation matrix $P \in
\mathcal{P}_d$ by solving the linear assignment problem via the
\textbf{Hungarian algorithm}:
\[
  P \;=\; \underset{P' \in \mathcal{P}_d}{\arg\min}
    \;\|S - P'\|_F^2,
\]
yielding a hard, discrete axis-aligned decomposition for inference. The
relaxed $S$ is used in place of $P$ only during training:
\[
  z \;=\; S h.
\]

\subsection{Datasets}
\label{subsec:datasets}

\begin{figure}[h!]
\centering
\includegraphics[width=\linewidth]{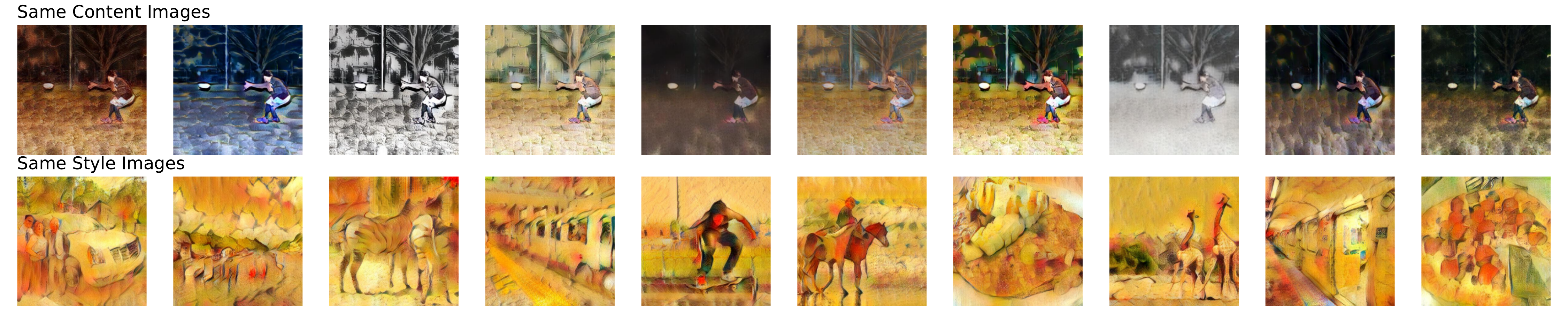}
\caption{Examples of images belonging to same content (Top), and same colour (Bottom)}
\label{fig:dataset_vision_examples}
\end{figure}

\begin{figure}[h!]
\centering
\includegraphics[width=\linewidth]{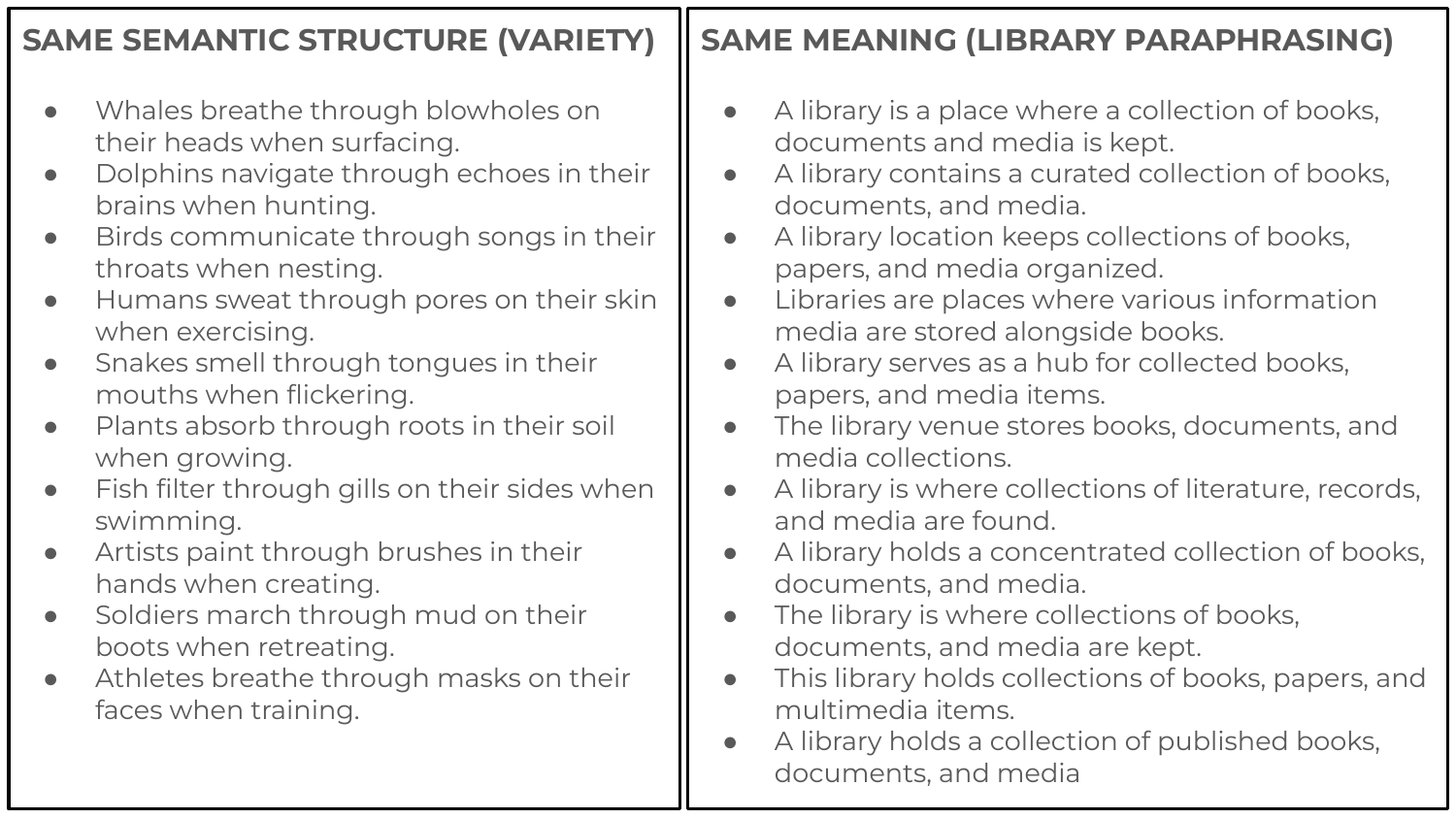}
\caption{Examples of sentences belonging to same syntax(top), and same semantics (bottom).}
\label{fig:lang_dataset_examples}
\end{figure}

\begin{figure}[h!]
\centering
\includegraphics[width=\linewidth]{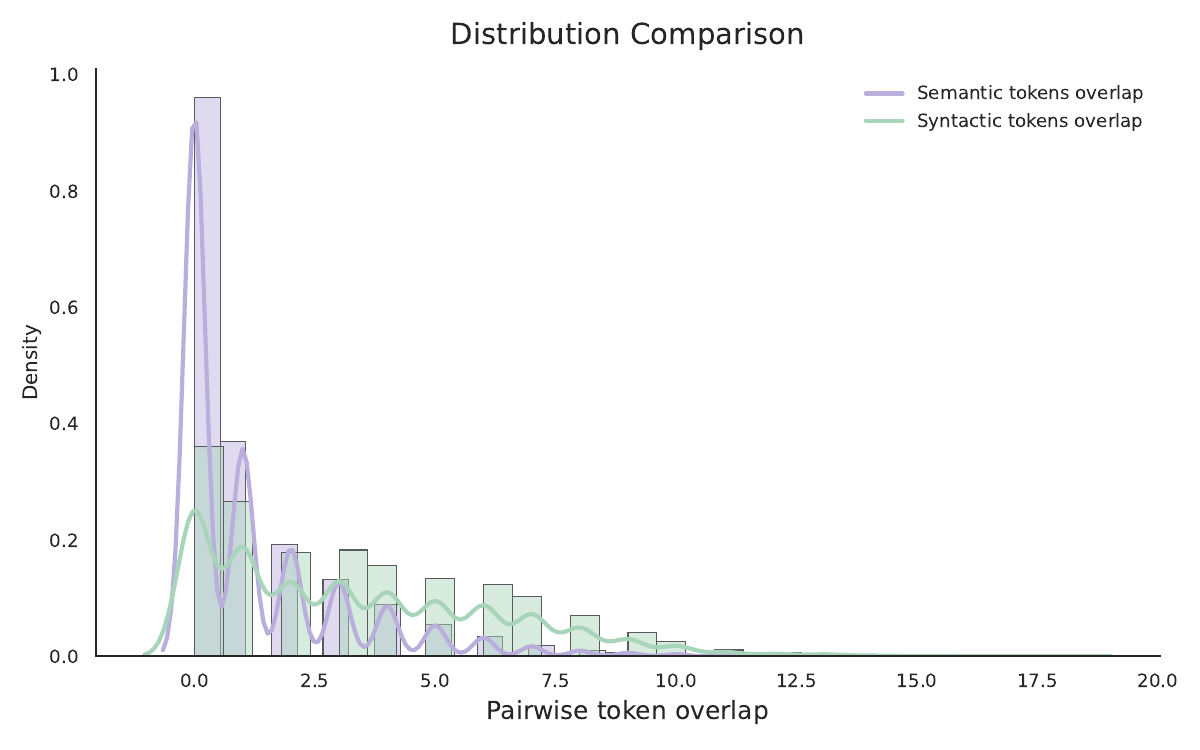}
\caption{For the semantic paraphrases and syntactic counterparts generated for each sentence in the Pereira dataset, we compare all sentence pairs within each semantic or syntactic set by measuring the number of tokens that remain identical at the same token positions across the two sentences. The resulting distributions reveal that semantic paraphrases exhibit a broader spread of token overlap values, reflecting greater variability in lexical structure, whereas syntactic counterparts are heavily concentrated near zero, indicating minimal token overlap at corresponding positions.}
\label{fig:token_overlap}
\end{figure}

\begin{table}[htbp]
  \centering
  \begin{tcolorbox}
    \centering
    \textbf{Prompt to generate paraphrases}\\[1ex]
    \begin{lstlisting}[basicstyle=\ttfamily\scriptsize,
                       breaklines=true,
                       columns=flexible,
                       frame=none]
Prompt:
f"""You are an expert image captioner. I'll show you some existing captions for an image, and your task is to generate 70 NEW captions that:
    1. Are similar in style and detail level to the existing captions
    2. Capture the same meaning but with different wording
    3. Are direct, concise descriptions (around 10-15 words each)
    4. Are worded differently from each existing caption and from each other

Here are the existing captions:
{insert all captions text for the image here}

Generate 10 new captions formatted exactly as:
1. [First new caption]
2. [Second new caption]
3. [Third new caption]
4. [Fourth new caption]
5. [Fifth new caption]
6. [Sixth new caption]
7. [Seventh new caption]
8. [Eighth new caption]
9. [Ninth new caption]
10. [Tenth new caption] ... """
    \end{lstlisting}
  \end{tcolorbox}
  \caption{Prompt used for generating paraphrase using Gemini-2.5-Flash.}
  \label{tab:para_prompt}
\end{table}

\begin{table}[htbp]
\centering
\begin{tcolorbox}
\centering
\textbf{Prompt to generate syntactic counterparts}
\begin{lstlisting}[basicstyle=\ttfamily\scriptsize,
breaklines=true,
columns=flexible,
frame=none]
Prompt:
f"""You are an expert meaningful syntactic sentence creator. I'll show a sentence, and your task is to generate 70 NEW syntactically similar sentences such that:
1. They retain the structure of the original sentence
2. They change some words of the original sentence to different words, and these new words should be a part of English vocabulary
3. The newly created sentences are syntactically similar to the original sentence
4. The newly created sentence should be meaningful

Here are the existing captions:
{insert all captions text for the image here}

Generate 10 new captions formatted exactly as:

1. [First new caption]
2. [Second new caption]
3. [Third new caption]
4. [Fourth new caption]
5. [Fifth new caption]
6. [Sixth new caption]
7. [Seventh new caption]
8. [Eighth new caption]
9. [Ninth new caption]
10. [Tenth new caption] ... """
    \end{lstlisting}
    \end{tcolorbox}
    \caption{Prompt used for generating syntactic counterparts using Gemini-2.5-Flash.}
    \label{tab:syn_para_prompt}
    \end{table}

\paragraph{Vision (content and style).}

In the visual domain, subspace $A$ corresponds to \textbf{content} and subspace $B$ to 
\textbf{style}. The subspace-$A$ counterpart $x^+_{\text{content}}$ is an image sharing 
the same object identity and spatial layout as $x$ but differing in surface 
appearance---e.g., a photograph of a cat paired with a stylized rendering of the same 
scene in an impressionist style. The subspace-$B$ counterpart $x^+_{\text{style}}$ is 
an image sharing the same texture, color palette, or artistic rendering as $x$ but 
depicting a different object or scene entirely.

To construct the training stimuli, we randomly sample 800 images from the MS-COCO 
dataset~\citep{lin2014microsoft} and 800 style images from the Paint by Numbers 
dataset~\citep{painter-by-numbers}. Each MS-COCO image is stylized using each of the 800 style 
images via the Stylized ImageNet procedure similar to \citep{geirhos2019}, yielding a 
training set of 640{,}000 image pairs. Separate sets of 100 content and 100 style images 
are held out for validation and test, producing 10{,}000 pairs each. Example in Appendix Figure \ref{fig:dataset_vision_examples}.

\paragraph{Language (semantics and syntax).}

In the linguistic domain, subspace $A$ corresponds to \textbf{semantics} and subspace 
$B$ to \textbf{syntax}. The subspace-$A$ counterpart $x^+_{\text{sem}}$ is a sentence 
conveying the same meaning as $x$ but expressed in a different syntactic 
form---e.g., an active-voice sentence paired with its passive-voice paraphrase. The 
subspace-$B$ counterpart $x^+_{\text{syn}}$ is a sentence sharing the same grammatical 
structure as $x$ but with different lexical content and meaning entirely.

We draw 3{,}900 anchor--paraphrase pairs from the MSR Paraphrase 
Corpus~\citep{dolan2005automatically}---a collection of naturally occurring sentence pairs 
judged to be semantically equivalent---using the anchor sentence as the positive semantic 
counterpart and generating one syntactic counterpart per anchor (a sentence sharing the 
same grammatical template but with different lexical content) via a prompted Gemini 2.5 flash 
language model~\citep{comanici2025gemini}. The full generation prompt is provided in 
Appendix Tables~\ref{tab:para_prompt},\ref{tab:syn_para_prompt}, and the quality of the generated pairs was verified through 
manual inspection.

For inference, we use the neuroimaging dataset of \citet{pereira2018toward}, comprising 
627 sentences presented to human subjects during fMRI recording. For each sentence in 
this dataset, we generate 70 semantic and 70 syntactic counterparts via the same 
procedure, yielding subspace neighborhoods used to extract disentangled representations 
at test time. Examples in Appendix Figure \ref{fig:lang_dataset_examples}.

\paragraph{Pereira 2018} - \cite{pereira2018toward}, who recorded brain responses from 16 native English speakers as they silently read short, syntactically and semantically diverse passages. The study comprised three experiments: (1) isolated words and simple phrases to probe basic lexical–semantic representations, (2) 384 full sentences grouped into 96 short narratives on varied concrete topics such as professions, clothing, birds, musical instruments, natural disasters, and crimes, and (3) an additional 243 full sentences organized into 72 narratives of similar thematic breadth. For our analyses, we focus on the five participants who completed both Experiments 2 and 3, using only these two experiments because they provide full-length sentence stimuli: 627 unique sentences in total, suitable for modeling rich compositional semantics. Paraphrases and Images are generated with respect to the original sentence similar to \citet{saha2025modeling}.

\paragraph{Continuous time recurrent networks }

We apply our modularity hierarchy to two well-studied RNN paradigms from computational neuroscience: the context-dependent integration task of \citet{mante2013context} and the multitask cognitive paradigm of \citet{yang2019task}. Both involve continuous-time recurrent networks (CTRNNs) with 256 units, and both have been central to debates about whether neural networks organize their computations modularly or distribute them across the full population. 

We evaluate decompositions through causal ablation: zeroing out each subspace during inference and measuring the accuracy drop on both tasks. A double dissociation requires that ablating subspace $A$ selectively impairs one task while ablating subspace $B$ selectively impairs the other. We quantify this with the double dissociation index:
\begin{equation}
    \text{DDI} = \max\!\Big(\min(\Delta^\text{sel}_1,\, \Delta^\text{sel}_2) - \max(\Delta^\text{col}_1,\, \Delta^\text{col}_2),\; 0\Big),
    \label{eq:ddi}
\end{equation}
where $\Delta^\text{sel}_i$ is the on-target accuracy drop for task $i$ and $\Delta^\text{col}_i$ is the off-target (collateral) drop. DDI is positive only when both ablations selectively impair their respective tasks more than they damage the other, and approaches 1 for perfect dissociations.

\paragraph{Context-dependent integration (Mante task).}
We trained CTRNNs on the task of \citet{mante2013context}, in which the network receives noisy motion and color evidence simultaneously and must report either the motion or color direction depending on a binary context signal (Figure~\ref{fig:mante}a,b). Networks were trained with tanh activations across 10 random seeds. At the unit level, individual neurons respond to mixtures of motion, color, and context \citep{mante2013context, rigotti2013importance}. We applied our decomposition hierarchy and evaluated the resulting subspaces through causal ablation: zeroing out the motion subspace during inference and measuring the impact on motion versus color decisions, and vice versa.

\paragraph{Multitask cognitive RNNs (Yang task).}

We next applied our hierarchy to networks trained on 11 cognitive tasks spanning sensorimotor mapping, perceptual decision-making, context-dependent integration, working memory, and categorical matching \citep{yang2019task}. We trained networks with both softplus and tanh activations. Following the taxonomy of \citet{yang2019task}, we evaluate three task pairs that span the spectrum from fully separable to fully shared computation:
 
\begin{itemize}
\item \textbf{Anti vs DM1 (disjoint):} Anti requires responding in the opposite direction of a single stimulus (sensorimotor remapping), while DM1 requires reporting which of two overlapping stimuli is stronger (perceptual evidence comparison). These tasks share no computational requirements.
\item \textbf{DM1 vs DlyDM1 (inclusive):} DlyDM1 performs the same comparison as DM1 but with a delay between stimuli, requiring additional working memory. DlyDM1 is computationally a superset of DM1.
\item \textbf{DM1 vs CtxDM1 (mixed):} CtxDM1 presents stimuli on both input modalities but must selectively attend to one based on a context cue, using DM1's comparison machinery plus context-dependent input gating. The two tasks share their core computation.
\end{itemize}


\subsection{Models}
\label{subsec:modeldetails}

\subsubsection{Vision Models}
\label{subsec:model_details}

\begin{enumerate}
    \item \textbf{ResNet family} \cite{he2016deep} (ResNet-50, ResNet-34, ResNet-18, ResNet-101):  
    Deep residual networks that use skip connections to mitigate vanishing gradients, enabling very deep convolutional architectures. 

    \item \textbf{ConvNeXt family} (ConvNeXt Tiny, Base, Large) \cite{liu2022convnet}:  
    A modernized CNN design that incorporates architectural ideas from Transformers, such as large kernels and inverted bottlenecks. 

    \item \textbf{Swin Transformers} \cite{liu2021swin} (Swin Small, Tiny, Base, Large):  
    Hierarchical vision Transformers that process images using shifted windows, providing linear computational complexity with respect to image size.

    \item \textbf{Vision Transformers (ViTs)} (ViT Small, Tiny, Base, Large) \cite{dosovitskiy2020image}:  
    Pure Transformer architectures that treat images as sequences of non-overlapping patches, capturing global context through self-attention.

    \item \textbf{DINO models} (DINO Small, Base) \cite{caron2021emerging}:  
    Self-supervised representations learned via knowledge distillation, producing strong, transferable visual features without labeled data and supporting both ViT and ResNet backbones.

    \item \textbf{VGGs} (VGG 11,13,16,19) \cite{simonyan2014very} - is characterized by its depth, consisting of 16 layers including 13 convolutional layers and 3 fully connected layers. VGG-16 is renowned for its simplicity and effectiveness as well as its ability to achieve strong performance on various computer vision tasks including image classification and object recognition.

\end{enumerate}

\subsubsection{Language Models}

\begin{enumerate}
   \item \textbf{GPT-2 XL} \cite{radford2019language}:  
    Decoder-only Transformer language models trained with next-token prediction, commonly used as baseline autoregressive LLMs across a range of model sizes.

    \item \textbf{Pythia 410M} \citet{biderman2023pythia} — from the Pythia model suite. This model was trained on approximately $2.9989 \times 10^{11}$ ($\approx300B$) tokens, corresponding to just under one epoch on the original Pile dataset, and approximately 1.5 epochs on the deduplicated Pile (which contains ~207B tokens per epoch). All models in the Pythia suite  are trained using mixed precision with FP16. For our experiments, we use the final checkpoint (step 143,000), i.e., the last available training step.

    \item \textbf{Pythia 160M} \citet{biderman2023pythia} — from the Pythia model suite. This model was trained on approximately $2.9989 \times 10^{11}$ ($\approx300B$) tokens, corresponding to just under one epoch on the original Pile dataset, and approximately 1.5 epochs on the deduplicated Pile (which contains ~207B tokens per epoch). All models in the Pythia suite are trained using mixed precision with FP16. For our experiments, we use the final checkpoint (step 143,000), i.e., the last available training step.
\end{enumerate}

\subsection{Experimental Details: Recurrent Neural Network Modularity}
\label{app:rnnnetworkappendix}

\subsubsection{Network architecture}

Both the Mante and Yang RNNs are continuous-time recurrent neural networks (CTRNNs) with the following dynamics:
\begin{equation}
    h_{t+1} = (1-\alpha)\, h_t + \alpha\, f\!\big(W_\text{rec}\, h_t + W_\text{in}\, u_t + b\big) + \xi_t,
    \label{eq:ctrnn}
\end{equation}
where $h_t \in \mathbb{R}^{256}$ is the hidden state, $u_t$ is the input, $f$ is the activation function, $\alpha = \Delta t / \tau$ is the integration constant with $\Delta t = 20\,\text{ms}$ and $\tau = 100\,\text{ms}$ (giving $\alpha = 0.2$), and $\xi_t \sim \mathcal{N}(0, \sigma_\text{noise}^2)$ is Gaussian noise injected during training. The noise scale is $\sqrt{2\alpha}\,\sigma$ to ensure the variance of the noise process is independent of the discretization step size.

\subsubsection{Mante context-dependent integration task}

\paragraph{Task structure.}
The network receives three scalar inputs at each timestep: motion coherence, color coherence, and a binary context signal ($+1$ = attend motion, $-1$ = attend color). Each trial consists of a fixation period (10 steps, 200\,ms) followed by a stimulus period (40 steps, 800\,ms), totaling 50 timesteps (1000\,ms). During the stimulus period, the motion and color inputs are the signed coherence values corrupted by Gaussian noise ($\sigma_\text{input} = 1.0$). Motion and color coherences are drawn uniformly from $\{-0.5, -0.25, -0.1, 0, 0.1, 0.25, 0.5\}$, and context is drawn as $\pm 1$ with equal probability.

The network produces two output logits at each timestep. The target is determined by the sign of the attended coherence: in the motion context, the network should indicate the direction of motion; in the color context, the direction of color. The decision is read out from the last 10 timesteps of the stimulus period.

\paragraph{Training.}
Networks were trained with tanh activation, 256 hidden units, and no recurrent bias. Recurrent weights were initialized as $W_\text{rec} \sim 0.1 \cdot \mathcal{N}(0, 1/n_\text{units})$. Training used Adam with learning rate $10^{-3}$, batch size 256, and cross-entropy loss over the last 10 timesteps. Networks were trained for 2000 epochs. Noise was injected into the hidden state during training ($\sigma_\text{hidden} = 0.05$) but not during evaluation. We trained 10 networks with different random seeds.

\paragraph{Performance.}
Trained networks achieve $85$--$90\%$ accuracy on non-zero coherence trials (trials where the attended coherence is nonzero and a correct answer exists). Performance is comparable across motion and color contexts and across seeds (Table~\ref{tab:mante_acc}).

\begin{table}[h]
\centering
\caption{Mante RNN accuracy on non-zero coherence trials (mean $\pm$ s.d. across 10 seeds).}
\label{tab:mante_acc}
\begin{tabular}{lcc}
\toprule
 & Motion context & Color context \\
\midrule
Accuracy & $87.5 \pm 1.2\%$ & $87.1 \pm 1.7\%$ \\
\bottomrule
\end{tabular}
\end{table}

\subsubsection{Yang multitask cognitive paradigm}

\paragraph{Task structure.}
Following \citet{yang2019task}, a single network is trained on 11 cognitive tasks simultaneously. The input consists of 76 dimensions: 1 fixation unit, two modality channels of 32 units each (ring-coded directional inputs), and 11 task identity units (one-hot encoding). The output consists of 33 dimensions: 1 fixation unit and 32 ring-coded directional outputs.

The 11 tasks are:
\begin{enumerate}
    \item \textbf{Go}: Report the direction of the stimulus.
    \item \textbf{Anti}: Report the opposite direction.
    \item \textbf{DM1}: Two stimuli on modality 1; report the stronger.
    \item \textbf{DM2}: Two stimuli on modality 2; report the stronger.
    \item \textbf{CtxDM1}: Stimuli on both modalities; attend to modality 1 only.
    \item \textbf{CtxDM2}: Stimuli on both modalities; attend to modality 2 only.
    \item \textbf{MultSenDM}: Stimuli on both modalities; integrate both to decide.
    \item \textbf{DlyDM1}: Like DM1, but stimuli are separated by a delay period.
    \item \textbf{DMS}: Delayed match-to-sample (respond if stimuli match).
    \item \textbf{DMC}: Delayed match-to-category (respond if stimuli share a category).
    \item \textbf{DNMC}: Delayed non-match-to-category (respond if categories differ).
\end{enumerate}

Stimuli are encoded as Gaussian bumps on the 32-unit ring with tuning width $\sigma = 1.0$ radian. The target output is a Gaussian bump centered on the correct response direction (for directional tasks) or a high fixation signal (for tasks requiring withholding a response). Trial timing follows \citet{yang2019task}, with a curriculum: short fixed timing for the first third of training, followed by paper-matched variable durations.

\paragraph{Training.}
Networks were trained with either softplus or tanh activation, 256 hidden units, and recurrent noise ($\sigma = 0.05$). Training used Adam with learning rate $10^{-3}$, batch size 64, MSE loss with sigmoid output activation, and loss masking during fixation periods. Networks were trained for 50{,}000 iterations, with tasks randomly interleaved. Input noise $\sigma_\text{input} = 0.01$ was added to all inputs.

\paragraph{Performance.}
Trained networks achieve $>95\%$ accuracy on all 11 tasks for both activation functions. Accuracy is measured as the fraction of trials where the decoded response direction (circular mean of output ring activity) falls within $\pi/5$ radians of the target direction, or where the fixation output is correctly maintained for withholding tasks.

\paragraph{Fractional task variance (FTV).}
Following \citet{yang2019task}, we compute the fractional task variance for each unit $i$ and task pair $(A, B)$ as:
\begin{equation}
    \text{FTV}_i(A, B) = \frac{\text{Var}_A(h_i) - \text{Var}_B(h_i)}{\text{Var}_A(h_i) + \text{Var}_B(h_i)},
\end{equation}
where $\text{Var}_A(h_i)$ is the variance of unit $i$'s activity across trials of task $A$ at the final stimulus timestep. FTV ranges from $-1$ (unit varies only for task $B$) to $+1$ (unit varies only for task $A$), with $0$ indicating equal variance for both tasks.

For the disjoint pair (Anti vs DM1), softplus networks yield $|\text{FTV}| = 0.74$ with 74\% of units classified as specialized ($|\text{FTV}| > 0.5$), while tanh networks yield $|\text{FTV}| = 0.39$ with only 30\% specialized. This replicates the finding of \citet{yang2019task} that tanh networks lack the clear cluster structure of softplus networks.

\subsubsection{Decomposition training}

\paragraph{Optimization.}
For each task pair and decomposition level, we train a decomposition module $\mathcal{D}$ (permutation, orthogonal, or linear) by minimizing a task-sufficiency objective. Given hidden states $\{h^A_i\}$ and $\{h^B_i\}$ from tasks $A$ and $B$ and their corresponding target outputs, $\mathcal{D}$ partitions the hidden state into subspaces $z_A$, $z_B$, and $z_\text{res}$. We train a readout head from each subspace and minimize the reconstruction loss for the corresponding task targets:
\begin{equation}
    \mathcal{L} = \mathcal{L}_\text{task}(f_A(z_A), y^A) + \mathcal{L}_\text{task}(f_B(z_B), y^B) + \lambda \cdot \mathcal{L}_\text{reg},
\end{equation}
where $f_A, f_B$ are linear readout heads, $y^A, y^B$ are task targets, and $\mathcal{L}_\text{reg}$ is a regularizer specific to each decomposition level (Sinkhorn entropy for permutations, Cayley parameterization for orthogonal matrices, condition number penalty for linear maps).

For all levels, we train for 2{,}000 epochs with Adam at learning rate $10^{-3}$.

\paragraph{Initialization.}
For the Yang multitask cognitive paradigm, decompositions are initialized using FTV rankings: units are sorted by FTV value, and the top-$d$ units define the initial subspace $A$, the bottom-$d$ define subspace $B$. For Sinkhorn, this initializes the doubly stochastic matrix $\log \alpha$. For orthogonal decomposition, the Cayley skew-symmetric matrix $S$ is initialized from the FTV-ranked permutation. For linear decomposition, we use a two-stage initialization: first train an orthogonal decomposition, then initialize the linear transformation matrix $T$ from the learned orthogonal matrix $Q$. 

\paragraph{Checkpointing.}
Because the optimization landscape is non-convex, training can sometimes degrade decomposition quality. We address this with a checkpoint strategy: every 100 epochs, we evaluate an offline DDI proxy (sum of absolute selective accuracy drops) and save the model state if it exceeds the previous best. The FTV-initialized state is also saved as a baseline. After training, we return whichever checkpoint achieved the highest DDI proxy. This guarantees that more flexible decomposition levels (orthogonal, linear) can never perform worse than their initialization from less flexible levels.

\subsubsection{Causal evaluation}

For each trained decomposition, we evaluate functional modularity by running the full RNN on held-out trials while intervening on the hidden state. Specifically, at each timestep we apply the decomposition $\mathcal{D}$ to the hidden state, zero out subspace $A$ (or $B$), reconstruct the hidden state from the remaining subspaces, and continue the forward pass. We then measure task accuracy under each ablation condition (none, ablate $A$, ablate $B$) on 500 held-out trials per task.

For the Yang task, accuracy is measured as the fraction of correct responses (decoded direction within $\pi/5$ of target). For the Mante task, accuracy is measured on non-zero coherence trials as the fraction of correct binary decisions (motion left/right or color left/right).

\subsubsection{Targeted dimensionality reduction (TDR) baseline}

For the Mante task, we additionally compute TDR axes \citep{mante2013context} as a baseline. For each task variable (motion coherence, color coherence, context, choice), we fit a linear regression from the hidden state at the final stimulus timestep to the scalar variable value across all trials. The regression coefficient vector (normalized to unit length) defines the TDR axis for that variable. We report $R^2$ values for each regression and the cosine similarity between axes.

To evaluate the causal effect of each axis, we project out the corresponding direction from the hidden state at every timestep during inference (replacing $h_t$ with $h_t - (h_t \cdot v) v$ where $v$ is the TDR axis) and measure task accuracy. We evaluate ablation of the motion axis, color axis, both axes, and the context axis, comparing accuracy to the intact baseline.

Across 10 seeds, TDR achieves $R^2 = 0.77 \pm 0.01$ for motion and $R^2 = 0.76 \pm 0.01$ for color, with near-orthogonal axes (cosine similarity $= -0.03 \pm 0.10$). The context axis is near-perfectly predictive ($R^2 > 0.999$). However, ablating any single axis changes accuracy by less than 0.5\%, indicating that these 1-dimensional axes are informationally relevant but not causally necessary (see Section~\ref{sec:rnn_results}).

\subsection{Experimental Details: Vision and Language model modularity}
\label{subsec:exptdetails}

We conduct all experiments using NVIDIA GeForce RTX 4090 GPUs. For training the orthogonal and invertible transformations, we use adam optimizer with a learning rate of $1 \times 10^{-5}$, while permutation-based transformations are trained with a learning rate of $1 \times 10^{-2}$. For the Sinkhorn relaxation, models are trained for 20 epochs. We use a batch size of 8 and apply gradient accumulation, updating model weights every 32 samples.

\subsection{Context-dependent integration (Mante task)}
\label{subsec:Mante task}

\begin{figure}[t]
    \centering
    \includegraphics[width=\linewidth]{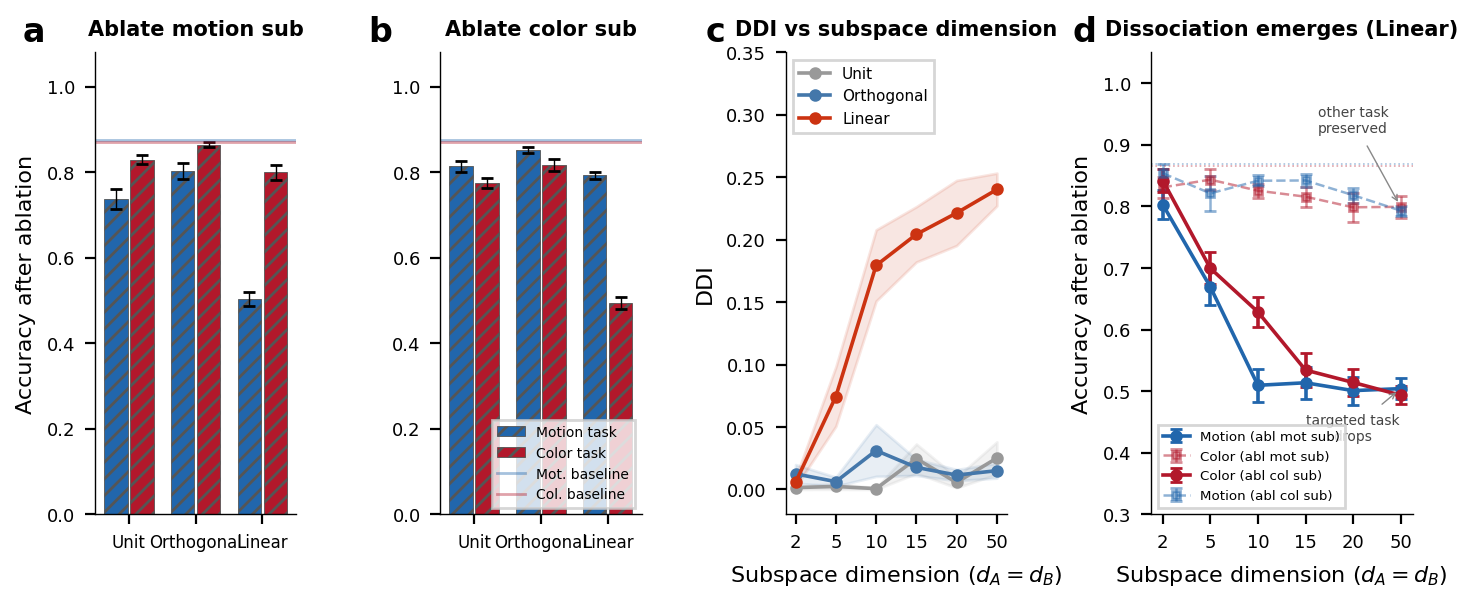}
    \caption{\textbf{a,b}~Accuracy on motion and color tasks after ablating the motion or color subspace (linear decomposition, $d=50$), shown for unit, orthogonal, and linear levels. Horizontal lines indicate intact baseline accuracy. Only linear decomposition produces selective impairment.
\textbf{(c)}~DDI as a function of subspace dimension for unit (gray), orthogonal (blue), and linear (red) decompositions. Unit and orthogonal DDI remain near zero at all dimensions; linear DDI rises to $\sim$0.28 by $d=15$.
\textbf{(d)}~Accuracy after ablation as a function of dimension (linear decomposition). Solid lines show the targeted task (which should drop); dashed lines show the spared task (which should remain high). The dissociation emerges progressively with increasing subspace dimension.}
    \label{fig:mante}
\end{figure}

We trained continuous time recurrent networks on the task of \citet{mante2013context}, in which the network receives noisy motion and color evidence simultaneously and must report either the motion or color direction depending on a binary context signal (Figure~\ref{fig:mante}a,b). Networks were trained with tanh activations across 10 random seeds. At the unit level, individual neurons respond to mixtures of motion, color, and context \citep{mante2013context, rigotti2013importance}. We applied our decomposition hierarchy and evaluated the resulting subspaces through causal ablation: zeroing out the motion subspace during inference and measuring the impact on motion versus color decisions, and vice versa.
 
Unit-level and orthogonal decompositions produce near-zero DDI across all subspace dimensions tested ($d_A = d_B \in \{2, 5, 10, 15, 20, 50\}$), confirming that no permutation or rotation of the representation reveals functionally separable components (Figure~\ref{fig:mante}c). In contrast, linear decomposition yields robust double dissociations. At $d = 15$, the linear decomposition achieves DDI $= 0.277 \pm 0.026$ (mean $\pm$ s.d. across seeds): ablating the motion subspace reduces motion accuracy by ${\sim}39\%$ while leaving color accuracy nearly intact (collateral $< 4\%$), and ablating the color subspace produces the symmetric pattern (Figure~\ref{fig:mante}a,b). The DDI increases with subspace dimension up to $d \approx 15$--$20$ and plateaus thereafter (Figure~\ref{fig:mante}c), indicating that the functionally relevant subspaces are low-dimensional relative to the full 256-dimensional representation.

As a baseline, we also applied targeted dimensionality reduction (TDR) \citep{mante2013context}, which identifies the single axis in representation space along which activity best predicts each task variable via linear regression. TDR successfully identifies motion and color axes with high predictive accuracy ($R^2 = 0.77 \pm 0.01$ and $0.76 \pm 0.01$ respectively) that are approximately orthogonal (cosine similarity $= -0.03 \pm 0.10$), consistent with the original findings of \citet{mante2013context}. However, ablating these axes during inference has negligible causal effect: accuracy drops by less than $0.5\%$ for both tasks across all 10 seeds. This dissociation between predictive and causal relevance indicates that while motion and color information is correlated with these single axes, the functionally necessary computation is distributed across a higher-dimensional subspace. Our linear decomposition, which identifies multi-dimensional subspaces optimized for task sufficiency, recovers the causal structure that single-axis methods miss. The contrast also highlights a key distinction: encoding information along a direction is not the same as depending on that direction for computation. 

\subsection{Evaluation of style and content subspaces on Silhoutte-Edge Stimuli}
\label{subsec:geirhos_silhoutte}

\begin{figure}[h!]
\centering
\includegraphics[width=0.9\linewidth]{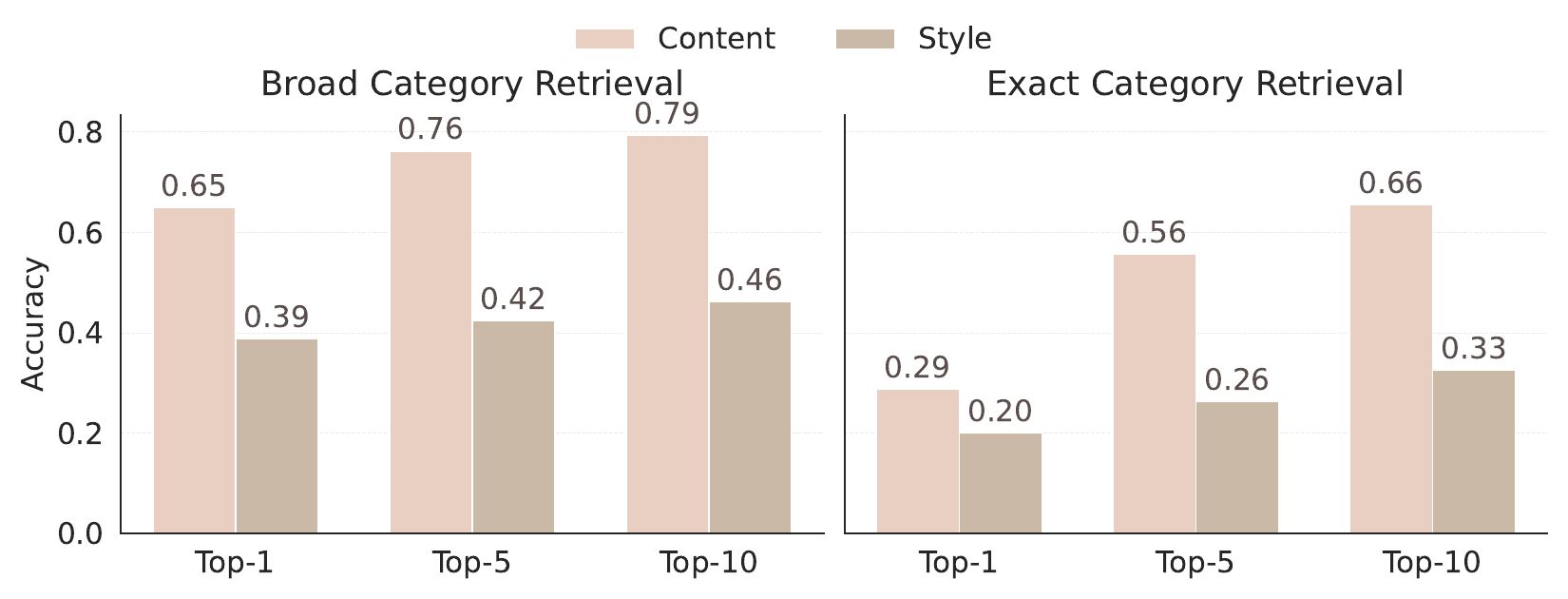}
\caption{Swin Base: Top-$K$ exact-category and broad-category retrieval for each silhouette image representation derived from edge-based inputs from \cite{geirhos2018imagenet}.}
\label{fig:geirhos_silhoutte_edge}
\end{figure}

We additionally evaluate our representations on the silhouette and edges dataset from 
\citet{geirhos2018imagenet}, which pairs each object silhouette with its corresponding 
edge map. For each silhouette image, we attempt to retrieve its correct edge-map 
counterpart using each subspace, under two retrieval criteria: \emph{instance-level} 
retrieval, where success requires recovering the edge map of the exact same image 
(e.g., the same bear photograph), and \emph{category-level} retrieval, where any edge 
map from the same object category suffices (e.g., any bear image). 

As expected, the content subspace substantially outperforms the style subspace on both 
criteria, consistent with the interpretation that shape and structural information are 
preferentially encoded in the content representation (Figure \ref{fig:geirhos_silhoutte_edge}, results for all models in Appendix Figure \ref{fig:exact_broad_all}). 

\subsection{Evaluation of style and content subspaces on Conflicting Cues}
\label{subsec:geirhos_conflicting_cues}

\begin{figure}[t]
    \centering
    \includegraphics[width=\linewidth]{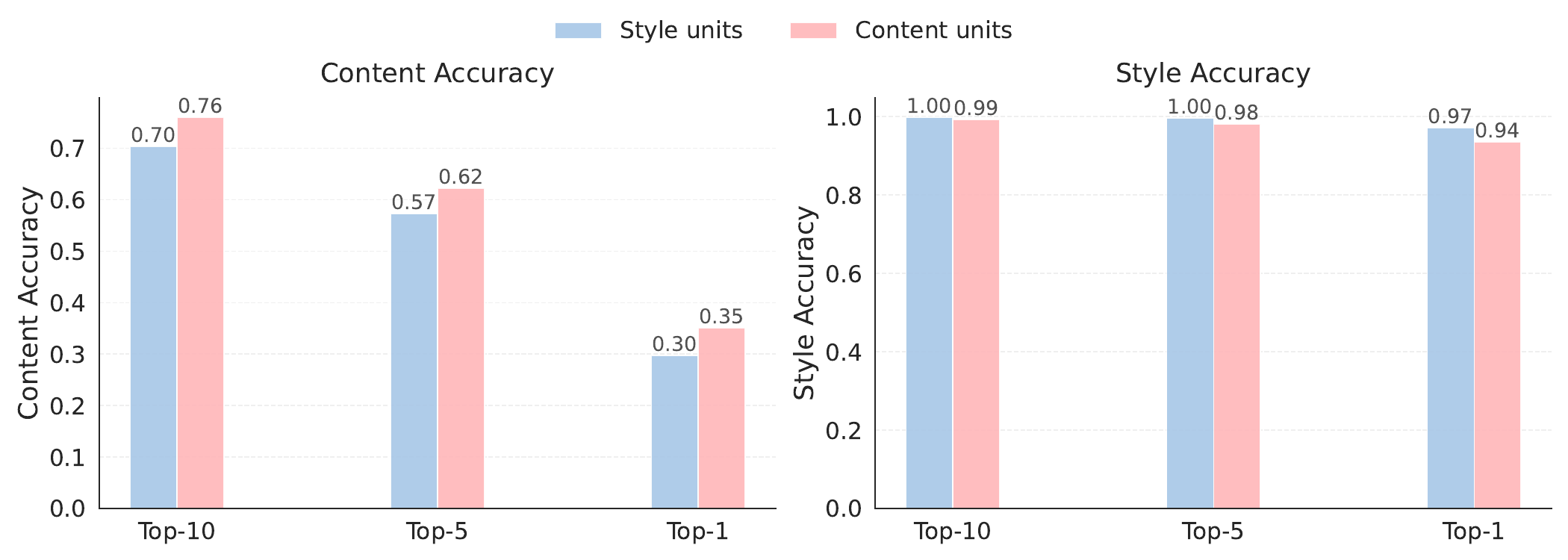}
    \caption{Top-$K$ content and style retrieval across content and style subspaces for the conflicting-cues images from \cite{geirhos2018imagenet}.}
    \label{fig:geirhos_conflicting_cues}
\end{figure}

We further validate our learned representations using the conflicting-cues stimuli introduced by \citet{geirhos2018imagenet}, which are explicitly designed to dissociate shape and texture information. This dataset consists of images in which the silhouette of one object category is filled with the texture of another---for example, the outline of a cat rendered with elephant skin---thereby placing content and style in direct opposition. Such stimuli provide a stringent test of subspace selectivity, as any residual entanglement between the two representations would lead to systematic retrieval errors.

We again perform a nearest-neighbor retrieval task using the content and style subspaces independently. For each query image, we assess whether the content subspace retrieves neighbors belonging to the same parent category defined by shape, and whether the style subspace retrieves neighbors sharing the same parent category defined by texture. The results (Figure~\ref{fig:geirhos_conflicting_cues}) are consistent with our earlier findings: the content subspace is selective for shape-based category membership, while the style subspace is selective for texture-based membership. This demonstrates that the learned decomposition remains robust even under adversarial combinations of content and style.

We note that style retrieval accuracy is uniformly high across both subspaces. This is partly due to the relative ease of the task, as multiple distinct textures are grouped under a shared higher-level category; consequently, retrieval does not require matching exact textures, but only coarse category-level consistency.

\subsection{Sensitivity to the Flexibility of the Learned Linear Map}
\label{subsec:ablation_transformation_flexibility}

A natural concern is whether the modular structure identified by our method reflects structure already present in the representation, or whether a sufficiently flexible linear transformation can simply reparameterize the representation to manufacture a task dissociation. This concern is particularly relevant for general invertible transformations: whereas orthogonal transformations preserve the geometry of the representation, general invertible maps can additionally rescale and shear different directions. We therefore performed a series of ablations and controls to test whether the flexibility of the learned transformation alone is sufficient to produce modularity. Across these experiments, we find that it is not: flexible linear reparameterization does not automatically produce a double dissociation, and successful dissociation depends on structure present in the underlying representation.

\paragraph{Failure cases under flexible transformations}

First, failure cases regarding presence of modular structure even after applying flexible transformations are already present in the Yang experiment. Appendix table \ref{tab:yang_ddi} reports the maximum DDI for each task-pair category across every tested linear subspace dimension and condition-number setting, thereby giving each category its most favorable result. Thus, even after optimizing the map and selecting the best linear setting, inclusive and mixed pairs remain nonmodular. This shows that optimization does not guarantee a double dissociation.

We additionally tested the optimization itself using causal interventions in 10 Mante RNNs (Appendix Table \ref{tab:mante_ddi}). For each learned map, we evaluated 100 random maps with exactly the same singular-value spectrum, thereby matching its full conditioning and distortion budget while randomizing only the orientation. We also refit five decompositions per network after independently shuffling the motion and color targets. None of the spectrum-matched random maps or maps refit with shuffled labels reached the learned DDI of its corresponding network. Thus, arbitrary maps with the same flexibility do not reproduce the learned causal split.

\begin{table}[t]
\centering
\caption{Maximum DDI values across Yang task-pair categories for softplus and tanh networks.}
\label{tab:yang_ddi}
\begin{tabular}{lcc}
\hline
\textbf{Yang task-pair category} &
\textbf{Softplus: maximum DDI} &
\textbf{Tanh: maximum DDI} \\
\hline
Disjoint pair   & 0.704 & 0.472 \\
Inclusive pair  & 0.052 & 0.000 \\
Mixed pair      & 0.048 & 0.002 \\
\hline
\end{tabular}
\end{table}

\begin{table}[t]
\centering
\caption{DDI values for different Mante decomposition conditions. Values are reported as mean $\pm$ SD across 10 networks.}
\label{tab:mante_ddi}
\begin{tabular}{lc}
\hline
\textbf{Mante decomposition} &
\textbf{DDI (mean $\pm$ SD across 10 networks)} \\
\hline
Learned map, true targets & $0.261 \pm 0.084$ \\
Spectrum-matched random maps & $0.0009 \pm 0.0004$ \\
Learned maps, shuffled targets & $0.010 \pm 0.008$ \\
\hline
\end{tabular}
\end{table}

\paragraph{Random-label control}

We randomly permute the content and style labels while training invertible transformations on SwinB representations. If our method learns an arbitrary projection in a high-dimensional space, then it should still be able to recover apparent content and style subspaces.
As shown in Appendix table \ref{tab:transformation_metrics}, this is not the case. Across all transformation classes the content/style metrics remain close to their chance values of 0 and selectivity scores remain near their chance value of 1. In contrast, the true content/style labels produce the strong dissociations reported in Table 1 in the main paper. 
This demonstrates that the invertible transformation is not capable of manufacturing modular structure for arbitrary label assignments. It succeeds when the supervision is consistent with structure already present in the representation.

\begin{table}[t]
\centering
\caption{Retrieval metrics and information-preservation measures for different transformations.}
\label{tab:transformation_metrics}
\small
\setlength{\tabcolsep}{4pt}
\begin{tabular}{lcccccccc}
\toprule
\textbf{Transformation} 
& \multicolumn{3}{c}{\textbf{Content Matching (CM)}} 
& \multicolumn{3}{c}{\textbf{Style Matching (SM)}} 
& \textbf{$R_s$} & \textbf{$R_c$} \\
\cmidrule(lr){2-4}
\cmidrule(lr){5-7}
& \textbf{Top 1} & \textbf{Top 5} & \textbf{Top 10}
& \textbf{Top 1} & \textbf{Top 5} & \textbf{Top 10}
& & \\
\midrule
Permutation 
& 0.0007 & 0.0023 & 0.0002 
& 0.0096 & 0.0154 & 0.0108 
& 1.0141 & 1.0011 \\

Orthogonal 
& 0.0034 & -0.0021 & 0.0014 
& -0.0024 & -0.0080 & 0.0036 
& 1.0275 & 1.0067 \\

Invertible 
& 0.0048 & 0.0065 & 0.0042 
& 0.0148 & -0.0034 & 0.0165 
& 0.9727 & 1.0286 \\
\bottomrule
\end{tabular}
\end{table}
\paragraph{Subspace-dimensionality control}

\begin{figure}[h!]
\centering
\includegraphics[width=0.9\linewidth]{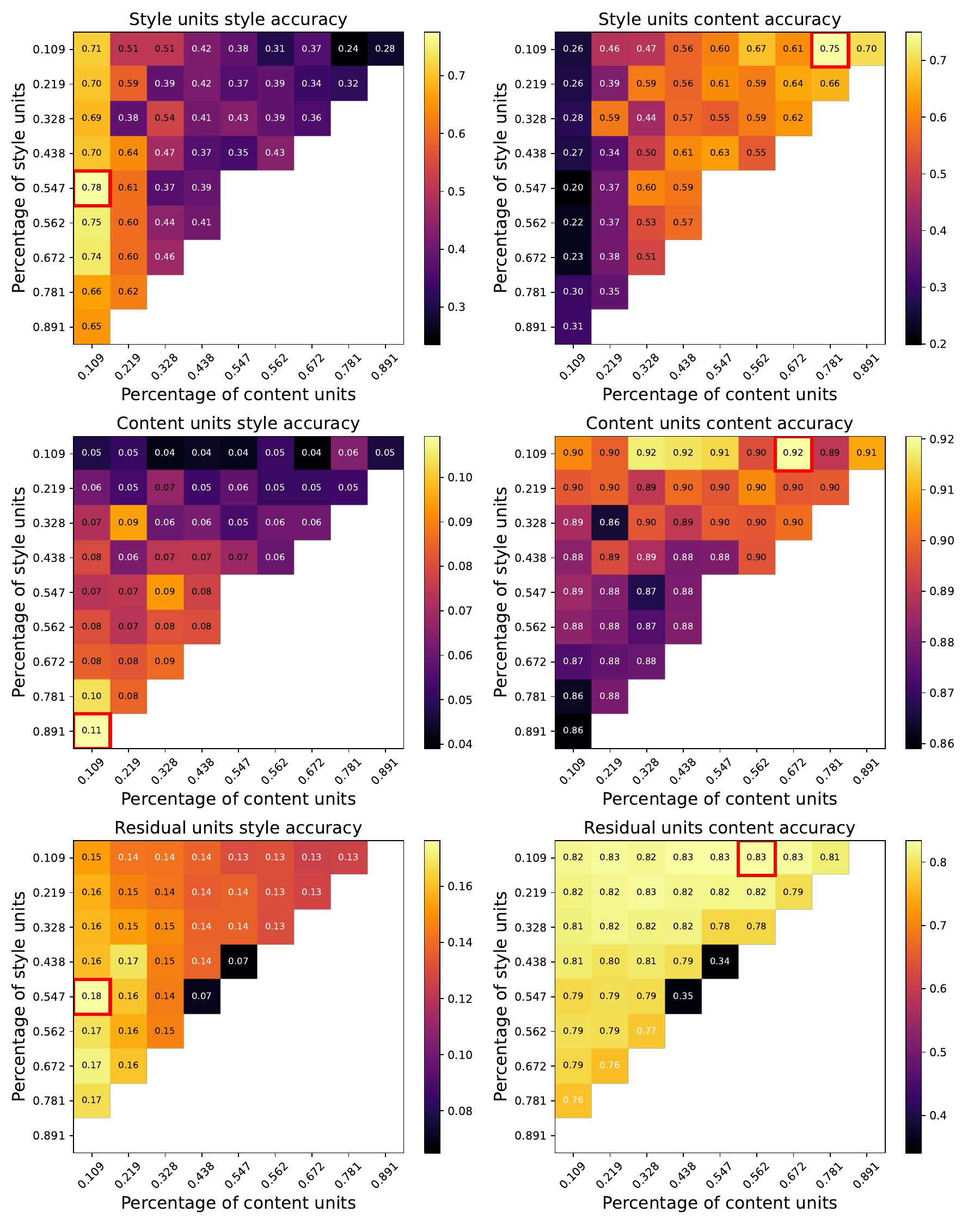}
\caption{For each subspace (style/content/residual) in visual representations obtained from Swin Base, we plot content and style selectivity with different portions of units allocated to content and style subspaces}
\label{fig:subspace_dim_sweep}
\end{figure}

We repeat the above experiments while varying the fraction of representation dimensions allocated to the content and style subspaces. Any remaining dimensions are assigned to the residual subspace. Appendix figure \ref{fig:subspace_dim_sweep} (fraction of style and content units in x and y axis) show that both content and style selectivity remain high across a broad range of allocations, provided that neither functional subspace is occupies more than approximately 40\% of the representation. 
We also audited the condition numbers across this sweep. While the learned invertible transformations are substantially more anisotropic than orthogonal transformations (which have condition number exactly one), the vast majority of solutions remained finite and well below values typically associated with numerical singularity. Some settings exhibited condition numbers on the order of $10^4$, indicating that the optimization occasionally exploits substantial anisotropic scaling. Importantly, successful double dissociation was observed across a wide range of condition numbers, suggesting that our conclusions are not tied to one favorable dimensional split or to a degenerate transform.

\paragraph{Intrinsically-overlapping function control}

We test whether our method can artificially produce modular subspaces even when the underlying functionalities are inherently non-modular. We constructed a Stylized-MNIST benchmark containing 10 digit identities rendered in 100 styles and trained the same Swin-B decomposition in two conditions: 
\begin{enumerate}
    \item \textbf{Content–Style Dissociation (Modular) - }the original setting used throughout the paper
Content–Category Dissociation
\item \textbf{(Non-Modular) - }instead of separating content from style, we attempt to separate image content from digit category. Here, images depicting the same digit are treated as belonging to the same category.
\end{enumerate}
Unlike content and style, content and category are not independent functionalities. Because exact content identifies its digit category, a content-sufficient subspace cannot simultaneously exclude category information. Consistent with this, applying both learned decompositions to the held-out set shows no double dissociation in this setting (when attempting to dissociate content from category, only the content metric remains high, while the category metric collapses) but replicates the content-style dissociation (Appendix table \ref{tab:dissociation_results}). 

\begin{table}[t]
\centering
\caption{Content dissociation results. Content matching (CM) measures preservation of content information, while the second set of metrics measures matching with the attribute being dissociated.}
\label{tab:dissociation_results}
\small
\setlength{\tabcolsep}{5pt}

\begin{tabular}{lccc|ccc}
\toprule
& \multicolumn{3}{c|}{\textbf{Content Matching (CM)}} 
& \multicolumn{3}{c}{\textbf{Style/Category Matching}} \\
\cmidrule(lr){2-4}
\cmidrule(lr){5-7}
\textbf{Dissociation}
& \textbf{Top 1} & \textbf{Top 5} & \textbf{Top 10}
& \textbf{Top 1} & \textbf{Top 5} & \textbf{Top 10} \\
\midrule

\textbf{Content--Style}
& 0.8157 & 0.8725 & 0.9351
& 0.5741 & 0.6635 & 0.7978 \\

\textbf{Content--Category}
& 0.8157 & 0.8725 & 0.9351
& -0.0800 & -0.1966 & -0.8368 \\

\bottomrule
\end{tabular}
\end{table}

One possible concern is that poor category dissociation is expected because successful content retrieval naturally implies successful category retrieval. To directly test whether the learned "category" and "content" subspaces actually encode different information, we apply both learned transformations to the stylized MSCOCO dataset used throughout the main paper and measure content retrieval accuracy from each subspace in Appendix table \ref{tab:dissociation_retrieval}.
For the content-style transformation, content retrieval is substantially better from the learned content subspace than from the style subspace, indicating that content information has indeed been selectively isolated. In contrast, for the content-category transformation, both subspaces achieve nearly identical content retrieval performance (the content subspace being only slightly better). In other words, content information remains distributed across both sets of units rather than becoming localized to a single subspace.

\begin{table}[t]
\centering
\caption{Content retrieval accuracy for content-style and content-category dissociation. 
Content units measure preservation of content information, while the second set of units 
corresponds to the attribute being dissociated.}
\label{tab:dissociation_retrieval}
\small
\setlength{\tabcolsep}{5pt}

\begin{tabular}{lccc|ccc}
\toprule
& \multicolumn{3}{c|}{\textbf{Content Units}} 
& \multicolumn{3}{c}{\textbf{Style/Category Units}} \\
\cmidrule(lr){2-4}
\cmidrule(lr){5-7}
\textbf{Dissociation}
& \textbf{Top 1} & \textbf{Top 5} & \textbf{Top 10}
& \textbf{Top 1} & \textbf{Top 5} & \textbf{Top 10} \\
\midrule

\textbf{Content--Style}
& 0.8424 & 0.9042 & 0.9269
& 0.7558 & 0.8450 & 0.8734 \\

\textbf{Content--Category}
& 0.8330 & 0.9021 & 0.9235
& 0.8186 & 0.8864 & 0.9105 \\

\bottomrule
\end{tabular}
\end{table}

Together, these results show that the learned transformation does not automatically produce modularity. The method has existing task-pair failure cases, random labels produce chance-level selectivity in the retrieval setting, and an intrinsically overlapping pair remains non-dissociable under the same training procedure that successfully separates content from style.

\subsection{What does the residual subspace contain?}
\label{subsec:app_residual_subspace}

The residual subspace can potentially contain three kinds of information: nuisance variables, computations shared by both tasks, or redundant task information not uniquely assigned to either functional subspace. We do not assume that it is uninformative. In the current objective, the residual receives no task-specific supervision and is retained only through the reconstruction loss. It is therefore best understood as the information-preserving remainder after the two targeted functional subspaces have been extracted.

To examine the residual empirically, we revisit the content-category dissociation experiment described above (Appendix Section~\ref{subsec:ablation_transformation_flexibility}). In this setting, the learned subspaces are optimized only to represent content and category, while style is never used as supervision. If the residual dimensions indeed preserve functionality outside the optimization objective, they should retain more style information than either the learned content or category subspaces.

Appendix table \ref{tab:style_dissociation} reports style retrieval accuracy from each of the three subspaces. The residual subspace achieves the highest style retrieval performance across all retrieval metrics. We also examined this directly in the Mante networks, where the two functions have clean behavioral readouts (Appendix table \ref{tab:dissocaiation_res_mante}). The motion and color subspaces have strongly selective effects, whereas ablating the residual reduces performance on both tasks by a similar amount.

\begin{table}[t]
\centering
\caption{Top-$k$ style retrieval accuracy for the Content--Category dissociation across content, category, and residual units.}
\label{tab:style_dissociation}
\small
\setlength{\tabcolsep}{4.5pt}
\renewcommand{\arraystretch}{1.1}
\begin{tabular}{lccc|ccc|ccc}
\toprule
\multirow{2}{*}{\textbf{Dissociation}}
& \multicolumn{3}{c|}{\textbf{Content}}
& \multicolumn{3}{c|}{\textbf{Category}}
& \multicolumn{3}{c}{\textbf{Residual}} \\
\cmidrule(lr){2-4}
\cmidrule(lr){5-7}
\cmidrule(l){8-10}
& \textbf{T1} & \textbf{T5} & \textbf{T10}
& \textbf{T1} & \textbf{T5} & \textbf{T10}
& \textbf{T1} & \textbf{T5} & \textbf{T10} \\
\midrule
Content--Category
& 0.141 & 0.258 & 0.331
& 0.152 & 0.275 & 0.346
& 0.194 & 0.352 & 0.449 \\
\bottomrule
\end{tabular}
\end{table}

\begin{table}[t]
\centering
\caption{Intervention-based ablation results for motion and color classification.}
\label{tab:dissocaiation_res_mante}
\small
\setlength{\tabcolsep}{8pt}
\renewcommand{\arraystretch}{1.15}
\begin{tabular}{lcc}
\toprule
\textbf{Intervention} & \textbf{Motion Accuracy} & \textbf{Color Accuracy} \\
\midrule
\textbf{Intact} 
& $\mathbf{0.872 \pm 0.016}$ 
& $\mathbf{0.872 \pm 0.022}$ \\

Ablate motion subspace 
& $0.508 \pm 0.072$ 
& $0.836 \pm 0.034$ \\

Ablate color subspace 
& $0.846 \pm 0.022$ 
& $0.558 \pm 0.083$ \\

Ablate residual 
& $0.724 \pm 0.050$ 
& $0.720 \pm 0.057$ \\
\bottomrule
\end{tabular}
\end{table}

 Thus, the residual is not purely nuisance: it contains or supports task-relevant information, but this information is shared or redundant rather than selectively associated with one task.

\begin{figure}[h!]
\centering
\includegraphics[width=0.9\linewidth]{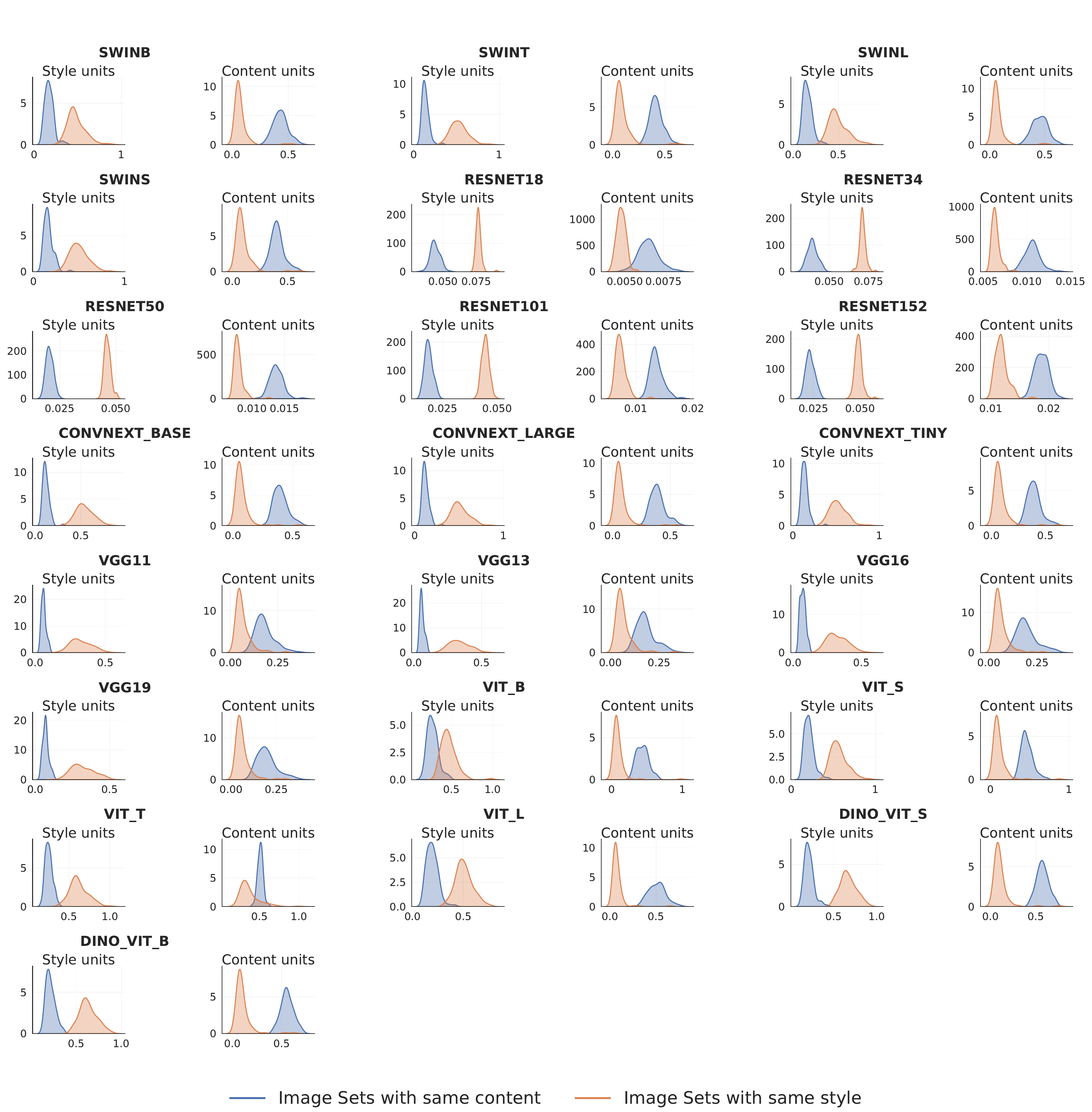}
\caption{For each subspace (style/content), we plot the distribution of average pairwise Pearson correlations computed within sets of images grouped by shared style or content, across invertible transforms for multiple vision models.}
\label{fig:distribution_invertible}
\end{figure}

\begin{figure}[h!]
\centering
\includegraphics[width=0.9\linewidth]{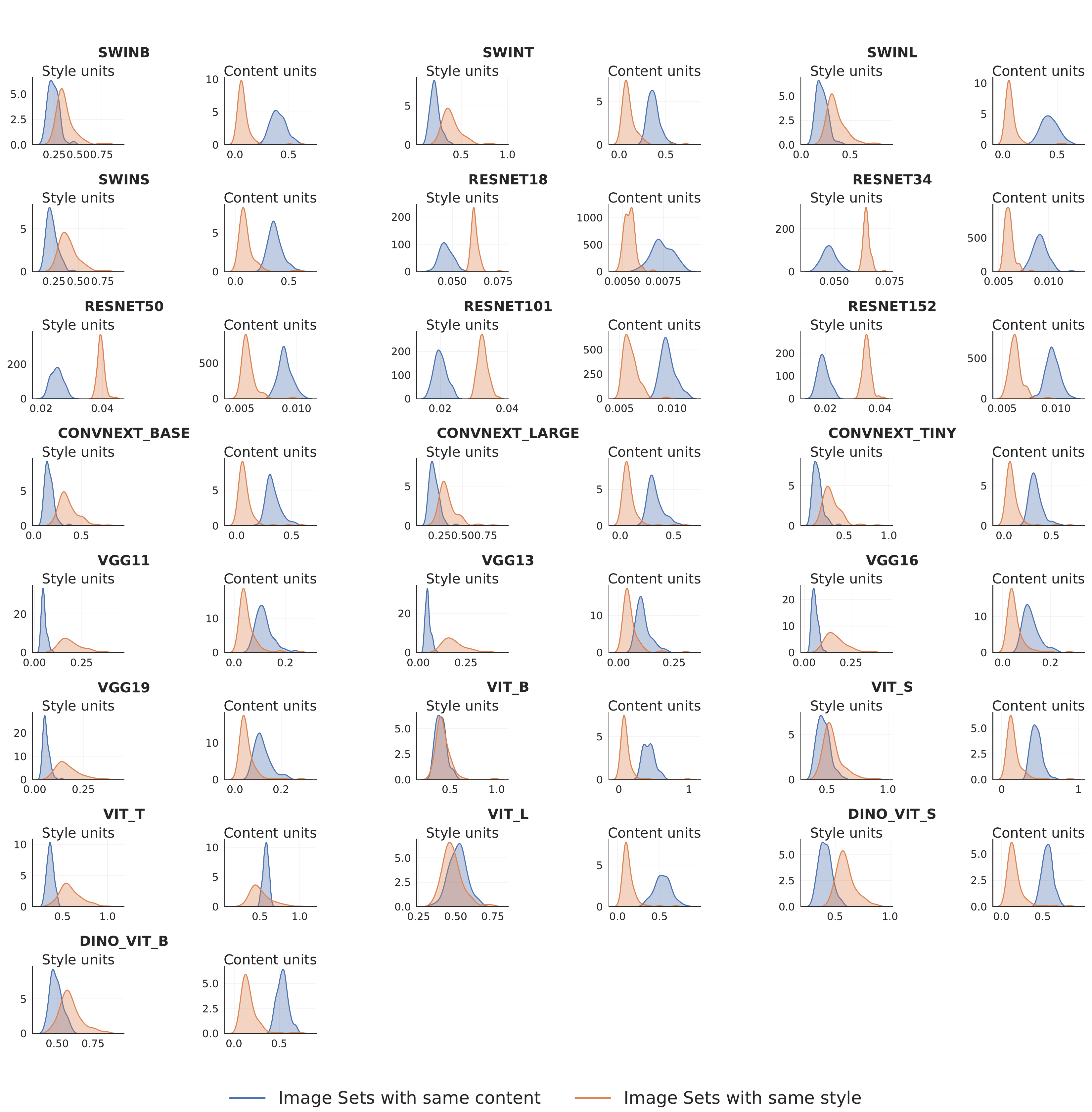}
\caption{For each subspace (style/content), we plot the distribution of average pairwise Pearson correlations computed within sets of images grouped by shared style or content, across orthogonal transforms for multiple vision models.}
\label{fig:distribution_orthogonal}
\end{figure}

\begin{figure}[h!]
\centering
\includegraphics[width=0.9\linewidth]{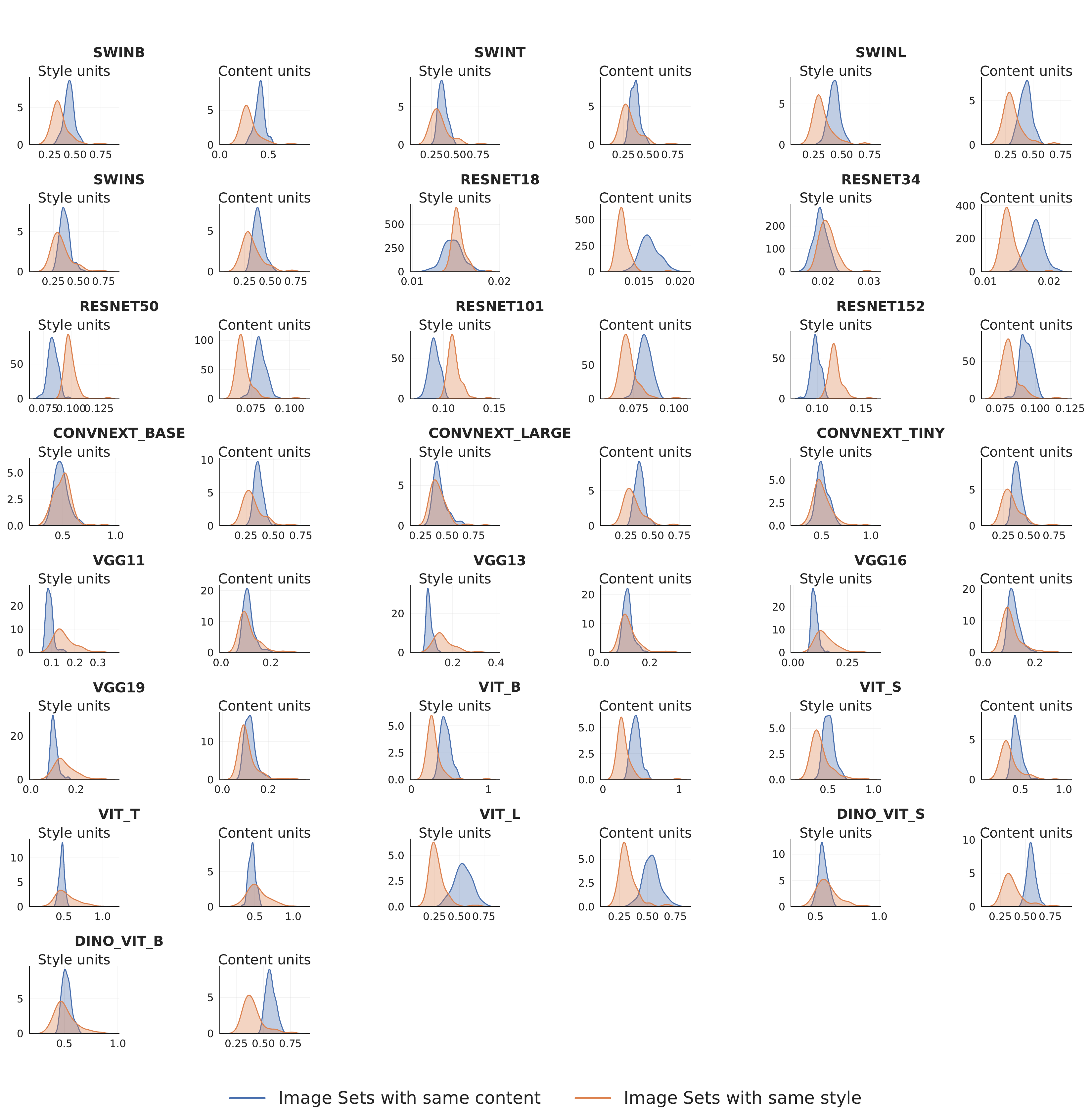}
\caption{For each subspace (style/content), we plot the distribution of average pairwise Pearson correlations computed within sets of images grouped by shared style or content, across permutation transforms for multiple vision models.}
\label{fig:distribution_permutation}
\end{figure}

\begin{figure}[h!]
\centering
\includegraphics[width=1.2\linewidth]{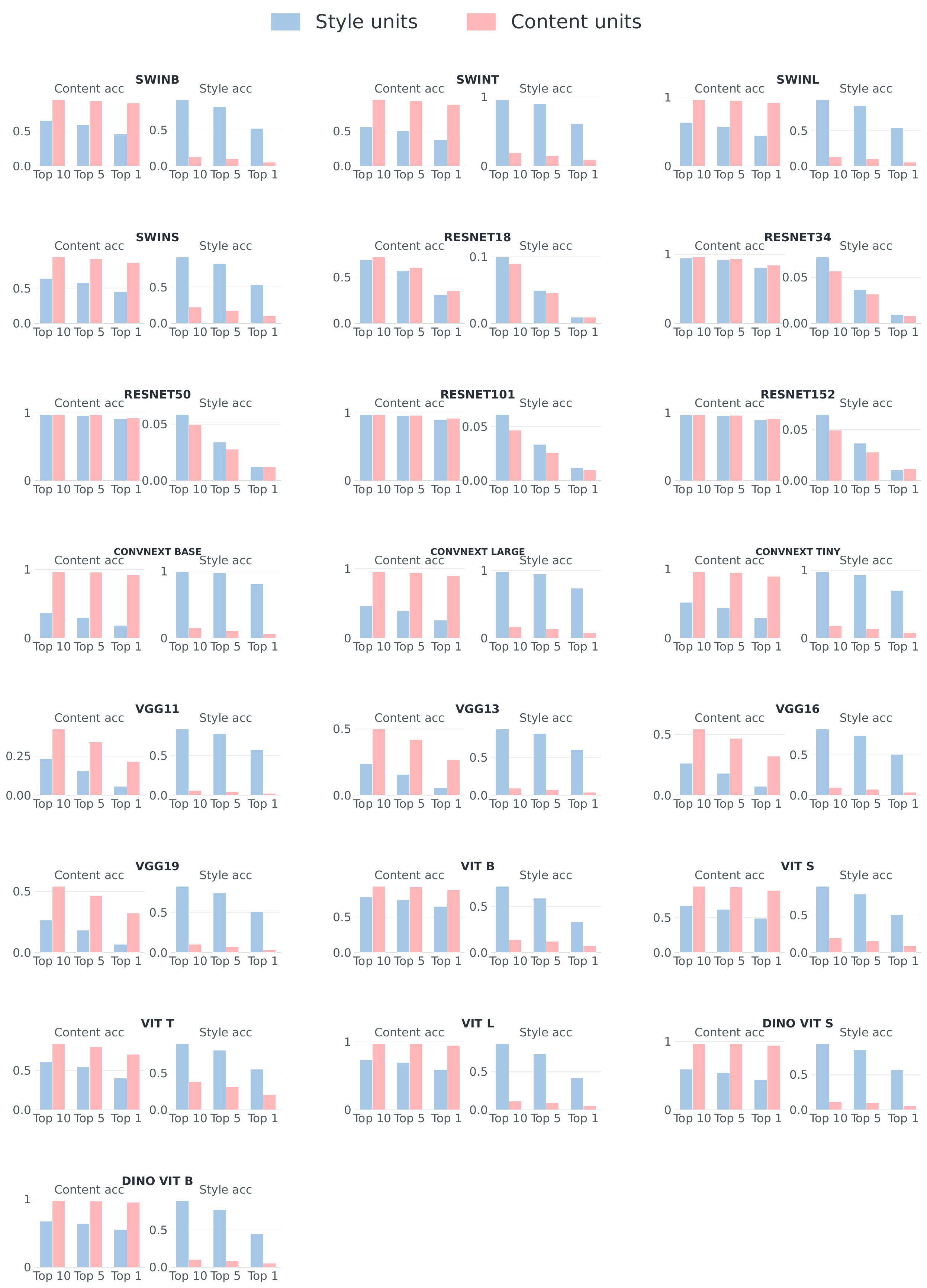}
\caption{Top-K content retrieval and Top-K style retrieval evaluated across content and style subspaces under invertible transforms for multiple vision models.}
\label{fig:retrieval_invertible}
\end{figure}

\begin{figure}[h!]
\centering
\includegraphics[width=1.2\linewidth]{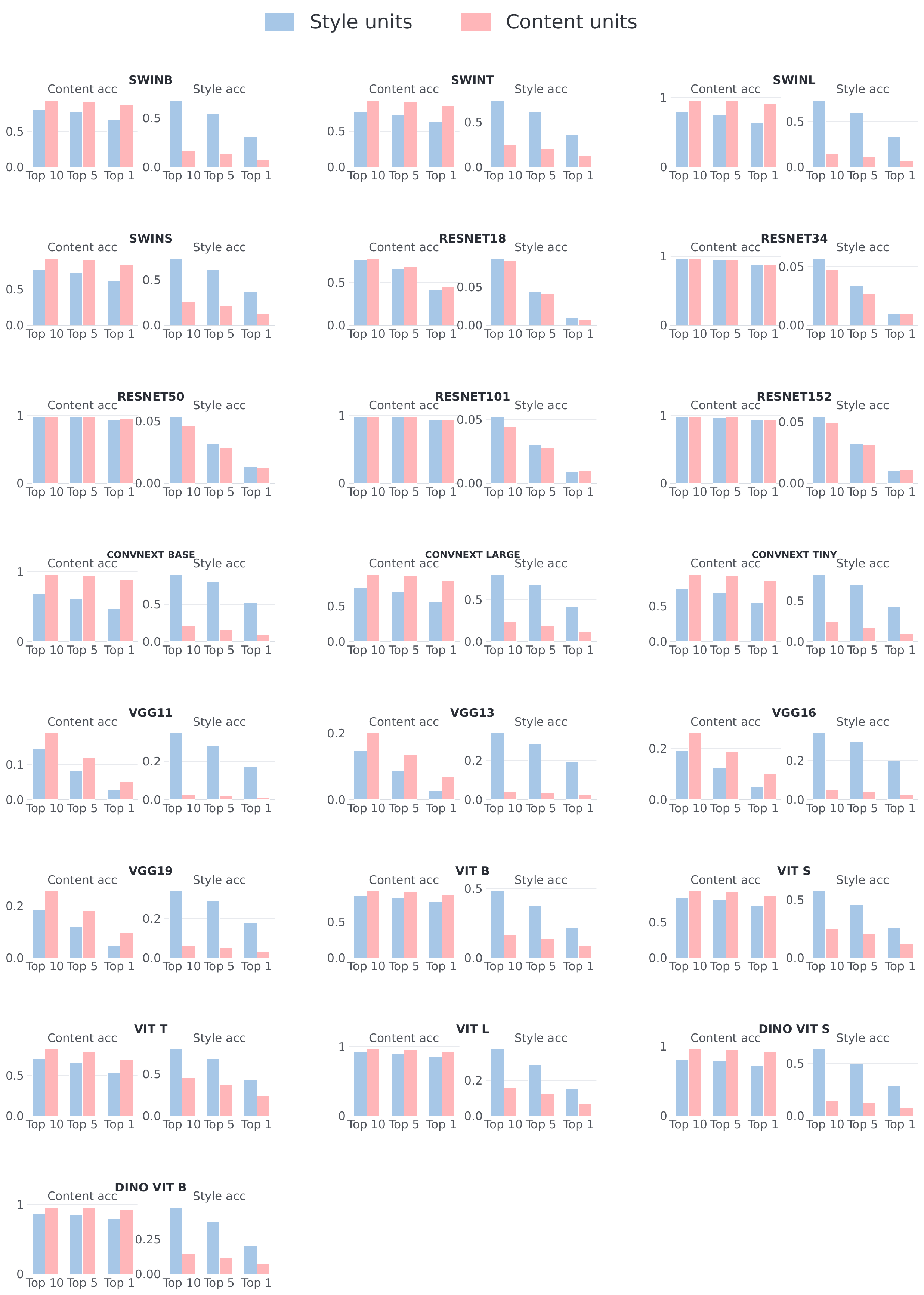}
\caption{Top-K content retrieval and Top-K style retrieval evaluated across content and style subspaces under orthogonal transforms for multiple vision models.}
\label{fig:retrieval_orthogonal}
\end{figure}

\begin{figure}[h!]
\centering
\includegraphics[width=1.2\linewidth]{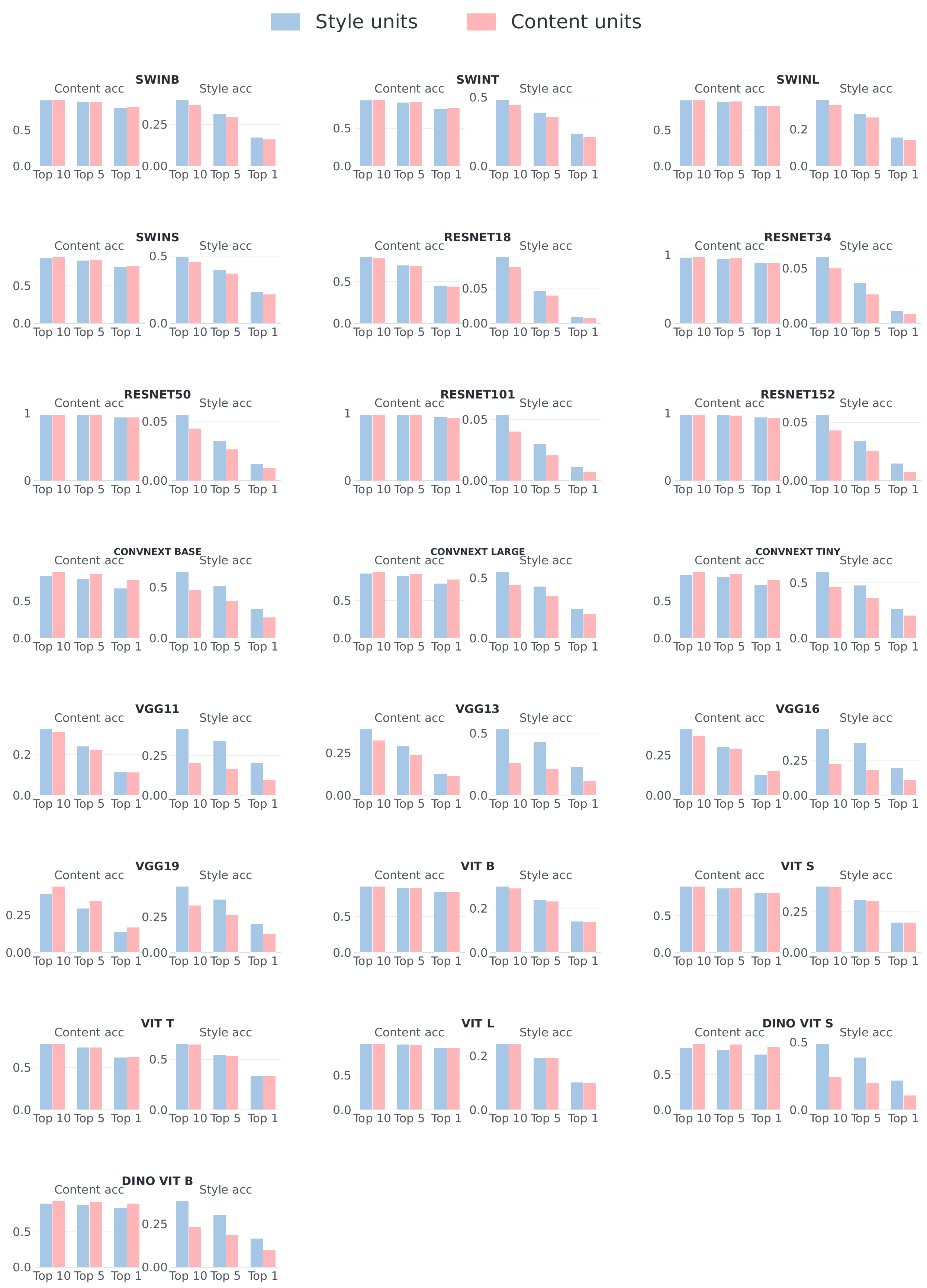}
\caption{Top-K content retrieval and Top-K style retrieval evaluated across content and style subspaces under permutation transforms for multiple vision models.}
\label{fig:retrieval_permutation}
\end{figure}

\begin{figure}[h!]
\centering
\includegraphics[width=\linewidth]{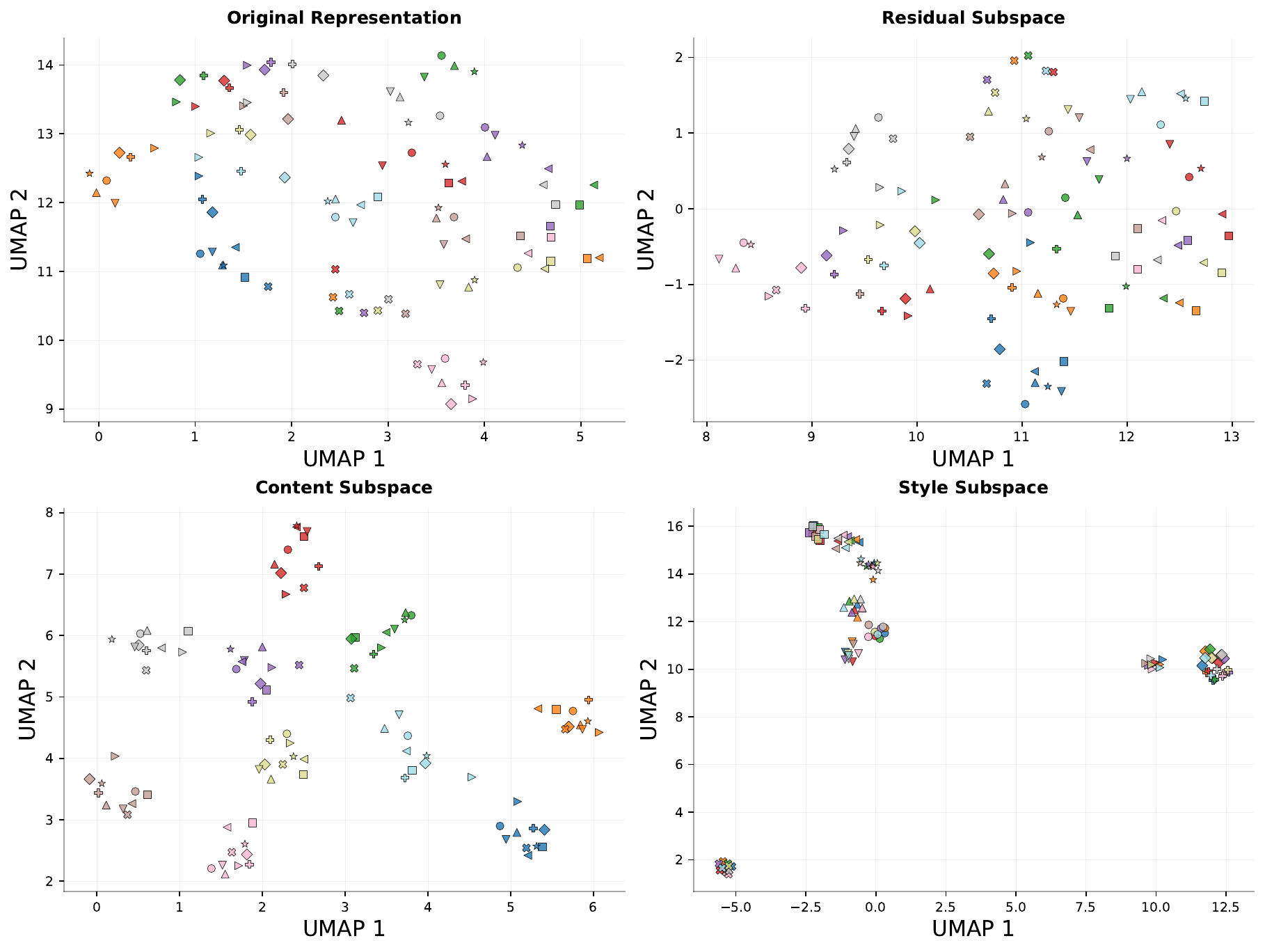}
\caption{We project the original visual representations from the punultimate layer of Swin Base, along with the dissociated content/style subspaces and residual subspaces, into 2D UMAPS to visualize how their manifolds are organized. Markers with the same color represent images with the same content, and markers with the same shape represent images with the same style.
We observe that images sharing the same functional attribute form coherent, well-defined clusters within the corresponding functional subspace manifold, indicating that the decomposition induces meaningful geometric organization aligned with the target attributes.
}
\label{fig:umap}
\end{figure}

\begin{table*}[ht]
\centering
\begin{minipage}[t]{0.48\textwidth}
\centering
\resizebox{\columnwidth}{!}{%
\begin{tabular}{@{}lrrrrrrrr@{}}
\toprule
\textbf{Transformation}
    & \textbf{Top-1 CM} $\uparrow$
    & \textbf{Top-5 CM} $\downarrow$
    & \textbf{Top-10 CM} $\uparrow$
    & \textbf{Top-1 SM} $\uparrow$
    & \textbf{Top-5 SM} $\uparrow$
    & \textbf{Top-10 SM} $\uparrow$
    & $R_s$ $\uparrow$
    & $R_c$ $\uparrow$ \\
\midrule
\multicolumn{9}{c}{\textit{Swin Base}} \\
\midrule
Permutation & 0.0063 & 0.0080 & 0.0125 & 0.0738 & 0.0625 & 0.0546 & 1.1032 & 1.0216 \\
Orthogonal  & 0.1381 & 0.1686 & 0.2435 & 0.7542 & 0.7550 & 0.7613 & 7.2348 & 2.5603 \\
Invertible  & 0.3107 & 0.3675 & 0.4884 & 0.8632 & 0.8785 & 0.8940 & 16.0815 & 5.2594 \\
\midrule
\multicolumn{9}{c}{\textit{Swin Small}} \\
\midrule
Permutation & 0.0138 & 0.0141 & 0.0180 & 0.0667 & 0.0633 & 0.0723 & 1.1016 & 1.0777 \\
Orthogonal  & 0.1714 & 0.1982 & 0.2696 & 0.6572 & 0.6565 & 0.6633 & 5.4447 & 2.5597 \\
Invertible  & 0.3224 & 0.3726 & 0.4778 & 0.7579 & 0.7842 & 0.8054 & 9.2309 & 4.8696 \\
\midrule
\multicolumn{9}{c}{\textit{Swin Tiny}} \\
\midrule
Permutation & 0.0048 & 0.0099 & 0.0177 & 0.0717 & 0.0746 & 0.0760 & 1.1899 & 1.0149 \\
Orthogonal  & 0.1716 & 0.2000 & 0.2591 & 0.6656 & 0.6606 & 0.6496 & 6.1425 & 2.4911 \\
Invertible  & 0.4076 & 0.4576 & 0.5709 & 0.8025 & 0.8294 & 0.8605 & 14.2392 & 6.2360 \\
\midrule
\multicolumn{9}{c}{\textit{Swin Large}} \\
\midrule
Permutation & 0.0045 & 0.0084 & 0.0091 & 0.0765 & 0.0717 & 0.0701 & 1.1161 & 1.0238 \\
Orthogonal  & 0.1682 & 0.2018 & 0.2889 & 0.7984 & 0.8019 & 0.8017 & 7.6299 & 3.2620 \\
Invertible  & 0.3421 & 0.3979 & 0.5173 & 0.8646 & 0.8830 & 0.9010 & 14.9196 & 6.2399 \\
\midrule
\multicolumn{9}{c}{\textit{Resnet 18}} \\
\midrule
Permutation & -0.0141 & -0.0160 & -0.0244 & 0.1490 & 0.1426 & 0.1209 & 1.0500 & 1.0524 \\
Orthogonal  & 0.0175  & 0.0275  & 0.0847  & 0.0379 & 0.0483 & 0.1720 & 1.0401 & 1.1309 \\
Invertible  & 0.0433  & 0.0607  & 0.1099  & 0.1047 & 0.0828 & 0.0430 & 1.0825 & 1.1680 \\
\midrule
\multicolumn{9}{c}{\textit{Resnet 34}} \\
\midrule
Permutation & 0.0038 & 0.0045 & 0.0017 & 0.1636 & 0.2747 & 0.2632 & 1.2109 & 1.0745 \\
Orthogonal  & 0.0037 & 0.0042 & 0.0083 & 0.1664 & 0.2141 & 0.0098 & 1.0608 & 1.0727 \\
Invertible  & 0.0137 & 0.0187 & 0.0367 & 0.2100 & 0.1315 & 0.1489 & 1.1855 & 1.1683 \\
\midrule
\multicolumn{9}{c}{\textit{Resnet 50}} \\
\midrule
Permutation & 0.0023  & 0.0033  & -0.0006 & 0.2115 & 0.2036 & 0.2394 & 1.2903 & 1.0444 \\
Orthogonal  & 0.0011  & 0.0023  & 0.0134  & 0.1439 & 0.0990 & 0.0227 & 1.1544 & 1.0842 \\
Invertible  & 0.0035  & 0.0097  & 0.0206  & 0.1624 & 0.1818 & 0.0244 & 1.1729 & 1.1735 \\
\midrule
\multicolumn{9}{c}{\textit{Resnet 101}} \\
\midrule
Permutation & 0.0006 & -0.0011 & -0.0067 & 0.2578 & 0.3180 & 0.3393 & 1.4054 & 1.0567 \\
Orthogonal  & 0.0026 & 0.0016 & 0.0041 & 0.1555 & 0.0671 & -0.0659 & 1.1270 & 1.0849 \\
Invertible  & 0.0048 & 0.0059 & 0.0179 & 0.2361 & 0.2292 & 0.1709 & 1.1688 & 1.1464 \\
\midrule
\multicolumn{9}{c}{\textit{Resnet 152}} \\
\midrule
Permutation & -0.0010 & -0.0043 & -0.0108 & 0.2381 & 0.2507 & 0.4694 & 0.0 & 0.0 \\
Orthogonal  & 0.0023  & 0.0028  & 0.0089  & 0.0944 & 0.0466 & -0.0667 & 1.1332 & 1.1204 \\
Invertible  & 0.0065  & 0.0070  & 0.0206  & 0.2380 & 0.2452 & -0.1262 & 1.2514 & 1.1383 \\
\midrule
\multicolumn{9}{c}{\textit{Convnext Tiny}} \\
\midrule
Permutation & 0.0401 & 0.0515 & 0.0919 & 0.2246 & 0.2319 & 0.2239 & 1.1119 & 1.3601 \\
Orthogonal  & 0.2168 & 0.2653 & 0.3637 & 0.7111 & 0.7513 & 0.7778 & 0.0 & 0.0 \\
Invertible  & 0.4612 & 0.5431 & 0.6754 & 0.8145 & 0.8568 & 0.8946 & 13.7342 & 6.3413 \\
\midrule
\multicolumn{9}{c}{\textit{Convnext Base}} \\
\midrule
Permutation & 0.0550 & 0.0764 & 0.1420 & 0.2704 & 0.2879 & 0.2821 & 0.9461 & 1.7463 \\
Orthogonal  & 0.2860 & 0.3504 & 0.4764 & 0.7641 & 0.8010 & 0.8211 & 7.3941 & 3.4337 \\
Invertible  & 0.6215 & 0.6890 & 0.8031 & 0.8470 & 0.8870 & 0.9251 & 15.2790 & 10.0803 \\
\bottomrule
\end{tabular}%
}
\end{minipage}
\hfill
\begin{minipage}[t]{0.48\textwidth}
\centering
\resizebox{\columnwidth}{!}{%
\begin{tabular}{@{}lrrrrrrrr@{}}
\toprule
\textbf{Transformation}
    & \textbf{Top-1 CM} $\uparrow$
    & \textbf{Top-5 CM} $\downarrow$
    & \textbf{Top-10 CM} $\uparrow$
    & \textbf{Top-1 SM} $\uparrow$
    & \textbf{Top-5 SM} $\uparrow$
    & \textbf{Top-10 SM} $\uparrow$
    & $R_s$ $\uparrow$
    & $R_c$ $\uparrow$ \\
\midrule
\multicolumn{9}{c}{\textit{Convnext Large}} \\
\midrule
Permutation & 0.0219 & 0.0332 & 0.0747 & 0.1948 & 0.1887 & 0.1716 & 0.9711 & 1.3686 \\
Orthogonal  & 0.1899 & 0.2335 & 0.3399 & 0.6986 & 0.7233 & 0.7216 & 5.1767 & 2.6059 \\
Invertible  & 0.5165 & 0.5888 & 0.7144 & 0.8328 & 0.8628 & 0.8936 & 12.9002 & 8.3816 \\
\midrule
\multicolumn{9}{c}{\textit{VGG 11}} \\
\midrule
Permutation & -0.0494 & -0.0690 & -0.0044 & 0.5089 & 0.5163 & 0.5239 & 1.7250 & 0.9651 \\
Orthogonal  & 0.2437  & 0.2934  & 0.4740  & 0.9346 & 0.9371 & 0.9273 & 5.2377 & 1.3887 \\
Invertible  & 0.4470  & 0.5480  & 0.7396  & 0.9261 & 0.9415 & 0.9568 & 14.0106 & 2.9038 \\
\midrule
\multicolumn{9}{c}{\textit{VGG 13}} \\
\midrule
Permutation & -0.2036 & -0.2185 & -0.1269 & 0.5038 & 0.5008 & 0.4889 & 1.8545 & 0.8998 \\
Orthogonal  & 0.2621  & 0.3606  & 0.6155  & 0.8846 & 0.8893 & 0.8853 & 4.6656 & 1.4102 \\
Invertible  & 0.5239  & 0.6298  & 0.7894  & 0.8912 & 0.9116 & 0.9383 & 13.7147 & 3.1449 \\
\midrule
\multicolumn{9}{c}{\textit{VGG 16}} \\
\midrule
Permutation & -0.1044 & -0.0393 & 0.1509 & 0.5285 & 0.5106 & 0.4446 & 2.1118 & 0.8849 \\
Orthogonal  & 0.2640  & 0.3407  & 0.5104 & 0.8522 & 0.8630 & 0.8705 & 3.9327 & 1.5968 \\
Invertible  & 0.5175  & 0.6118  & 0.7740 & 0.8836 & 0.9034 & 0.9253 & 11.0569 & 3.7754 \\
\midrule
\multicolumn{9}{c}{\textit{VGG 19}} \\
\midrule
Permutation & 0.1116 & 0.1428 & 0.1633 & 0.2840 & 0.2954 & 0.3457 & 1.3559 & 1.1274 \\
Orthogonal  & 0.2766 & 0.3537 & 0.5359 & 0.8170 & 0.8271 & 0.8210 & 3.4779 & 1.6213 \\
Invertible  & 0.5094 & 0.6094 & 0.7907 & 0.8768 & 0.8994 & 0.9212 & 9.8997 & 3.7337 \\
\midrule
\multicolumn{9}{c}{\textit{ViT Small}} \\
\midrule
Permutation & -0.0009 & 0.0021 & 0.0051 & 0.0082 & 0.0164 & 0.0060 & 0.9214 & 1.0033 \\
Orthogonal  & 0.0949 & 0.1091 & 0.1500 & 0.5705 & 0.5529 & 0.5233 & 3.6250 & 1.8224 \\
Invertible  & 0.2938 & 0.3416 & 0.4477 & 0.7818 & 0.8003 & 0.8174 & 10.0129 & 4.9636 \\
\midrule
\multicolumn{9}{c}{\textit{ViT Large}} \\
\midrule
Permutation & -0.0024 & -0.0032 & 0.0018 & 0.0041 & 0.0073 & 0.0049 & 0.8765 & 0.9984 \\
Orthogonal  & 0.0456 & 0.0559 & 0.0816 & 0.5709 & 0.5609 & 0.5383 & 4.2343 & 1.4566 \\
Invertible  & 0.2468 & 0.2827 & 0.3776 & 0.8694 & 0.8793 & 0.8785 & 17.0332 & 5.3963 \\
\midrule
\multicolumn{9}{c}{\textit{ViT Tiny}} \\
\midrule
Permutation & 0.0023 & -0.0033 & 0.0076 & 0.0111 & 0.0212 & 0.0085 & 0.9899 & 1.0039 \\
Orthogonal  & 0.1445 & 0.1642 & 0.2343 & 0.4293 & 0.4498 & 0.4414 & 2.3615 & 1.8058 \\
Invertible  & 0.2753 & 0.3234 & 0.4257 & 0.5777 & 0.6100 & 0.6258 & 4.0475 & 3.0856 \\
\midrule
\multicolumn{9}{c}{\textit{ViT Base}} \\
\midrule
Permutation & 0.0012 & -0.0004 & 0.0052 & 0.0260 & 0.0231 & 0.0206 & 0.9789 & 1.0307 \\
Orthogonal  & 0.0661 & 0.0858 & 0.1195 & 0.6628 & 0.6377 & 0.6000 & 6.3848 & 1.6223 \\
Invertible  & 0.1634 & 0.1927 & 0.2690 & 0.8022 & 0.7958 & 0.7761 & 11.8301 & 2.7591 \\
\midrule
\multicolumn{9}{c}{\textit{DINO ViT Base}} \\
\midrule
Permutation & 0.0355 & 0.0444 & 0.0726 & 0.3908 & 0.3779 & 0.3941 & 1.7156 & 1.4055 \\
Orthogonal  & 0.0928 & 0.1042 & 0.1387 & 0.6946 & 0.6760 & 0.6424 & 5.1234 & 2.2746 \\
Invertible  & 0.3064 & 0.3421 & 0.4194 & 0.8870 & 0.8927 & 0.8885 & 14.0465 & 7.6564 \\
\midrule
\multicolumn{9}{c}{\textit{DINO ViT Small}} \\
\midrule
Permutation & 0.0670 & 0.0860 & 0.1220 & 0.4987 & 0.4871 & 0.5032 & 2.1401 & 1.6032 \\
Orthogonal  & 0.1499 & 0.1717 & 0.2234 & 0.7660 & 0.7506 & 0.7306 & 6.4548 & 2.9013 \\
Invertible  & 0.3851 & 0.4304 & 0.5337 & 0.8770 & 0.8899 & 0.9020 & 14.2497 & 8.8929 \\
\bottomrule
\end{tabular}%
}
\end{minipage}
\caption{Comparison of Top-K content and style summary metrics, along with content and style selectivity ratios, across permutation, orthogonal, and invertible transformations across different vision models}
\end{table*}

\begin{figure}[h!]
\centering
\includegraphics[width=0.9\linewidth]{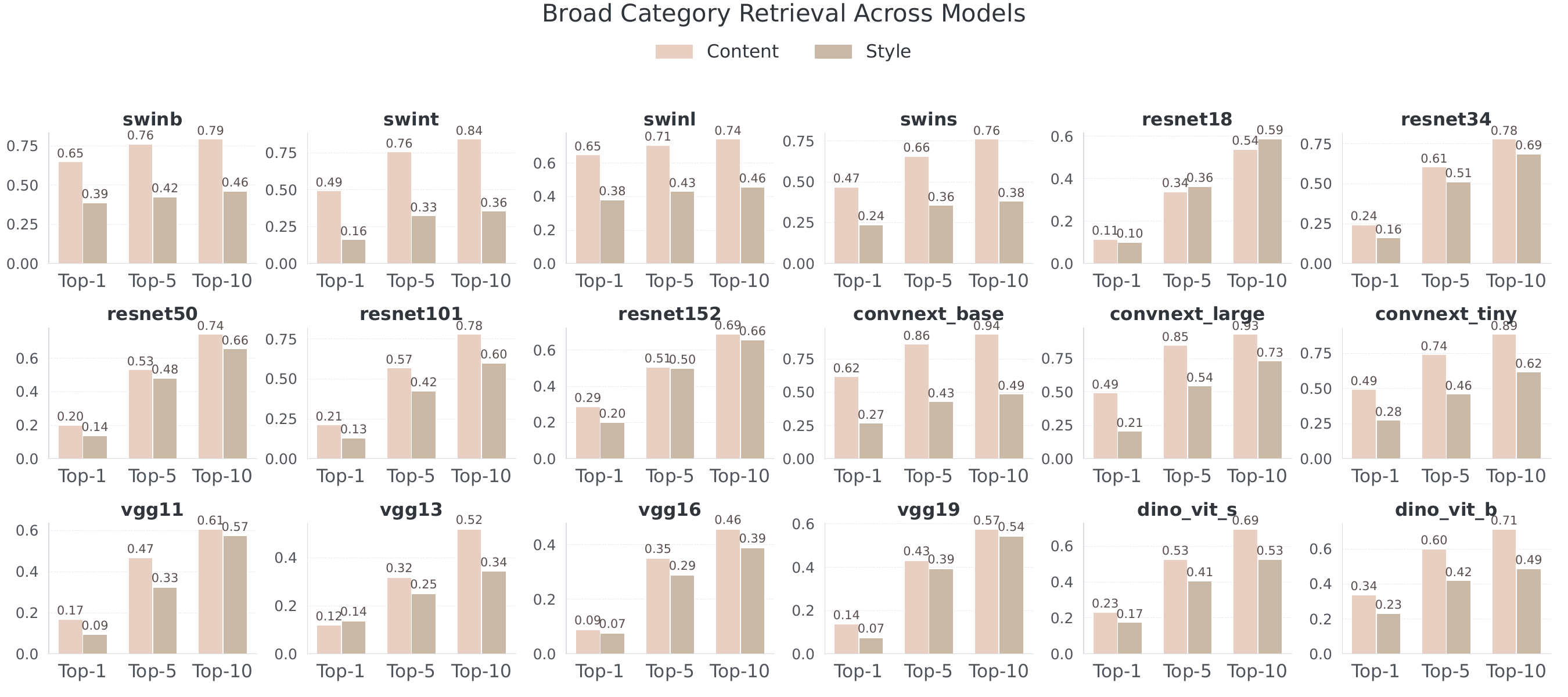}
\caption{Top-$K$ exact-category and broad-category retrieval for each silhouette image representation derived from edge-based inputs from \cite{geirhos2018imagenet} across multiple vision models under invertible transforms}
\label{fig:exact_broad_all}
\end{figure}

\begin{figure}[h!]
\centering
\includegraphics[width=0.9\linewidth]{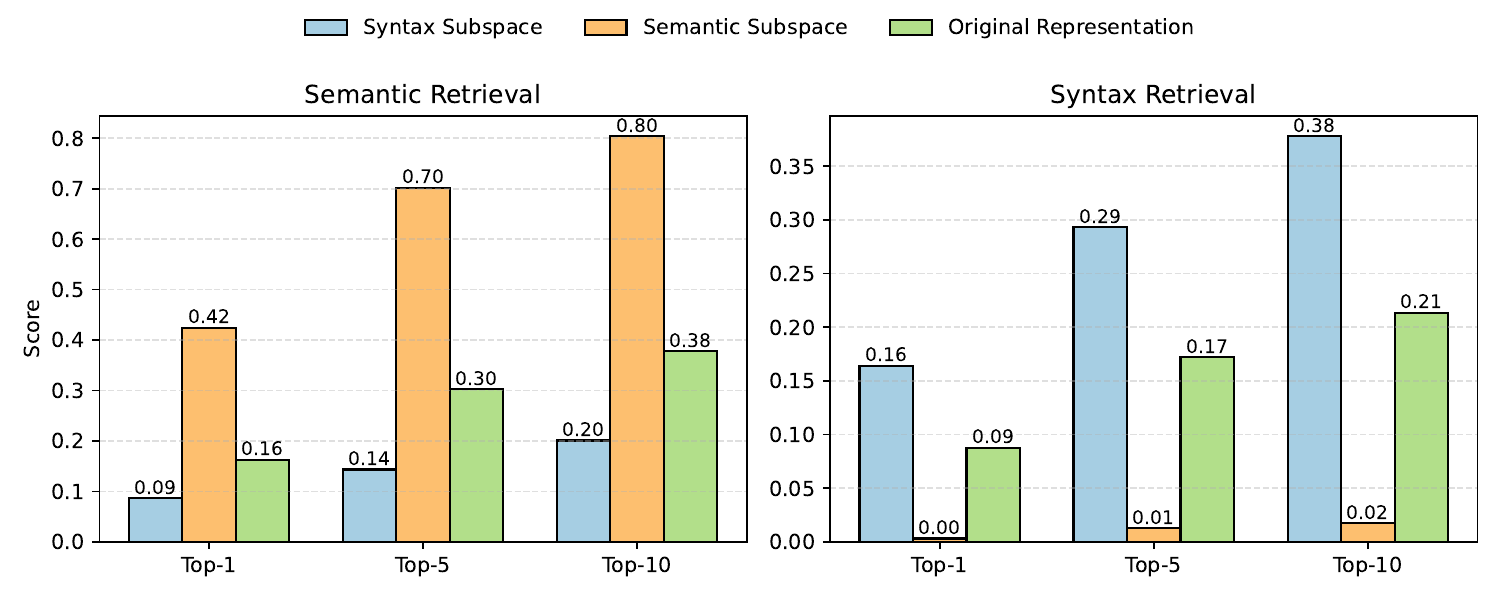}
\caption{(A) Top-K Semantic retrieval; (B) Top-K Syntax retrieval for Pythia 160M obtained using invertible transforms}
\label{fig:syn_sem_pythia_160m}
\end{figure}

\begin{figure}[h!]
\centering
\includegraphics[width=0.9\linewidth]{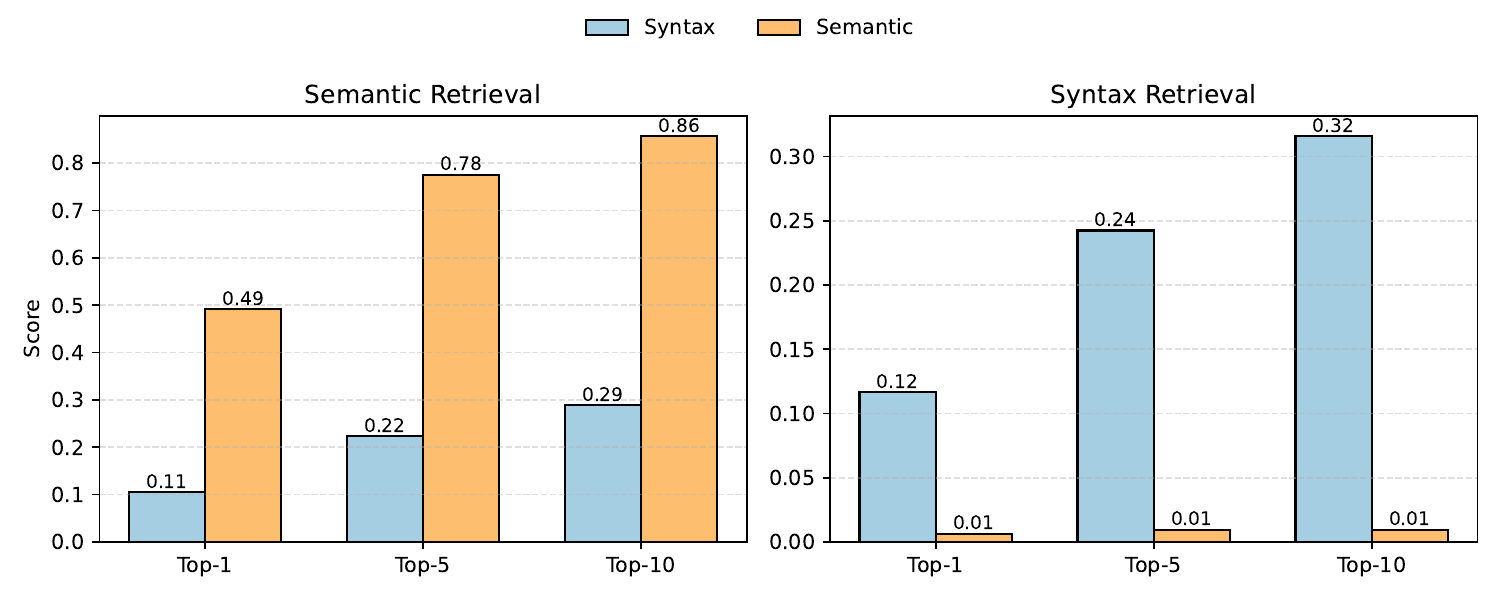}
\caption{(A) Top-K Semantic retrieval; (B) Top-K Syntax retrieval for Pythia GPT2-XL obtained using invertible transforms}
\label{fig:syn_sem_gpt2xl}
\end{figure}

\begin{table}[ht]
\centering
\resizebox{\columnwidth}{!}{%
\begin{tabular}{@{}lrrrrrrrr@{}}
\toprule
\textbf{Transformation}
    & \textbf{Top-1 $S_1$} $\uparrow$
    & \textbf{Top-5 $S_1$} $\uparrow$
    & \textbf{Top-10  $S_1$} $\uparrow$
    & \textbf{Top-1  $S_2$} $\uparrow$
    & \textbf{Top-5  $S_2$} $\uparrow$
    & \textbf{Top-10 $S_2$} $\uparrow$ \\
\midrule
\multicolumn{6}{c}{\textit{Pythia 160M}} \\
\midrule
Permutation & 0.1707  & 0.1418  & 0.1271  & 0.0357  & 0.0563  & 0.0861 \\
Orthogonal  & 0.9540  & 0.9058  & 0.8580  & 0.5574  & 0.5460  & 0.5259 \\
Invertible  & 0.9806  & 0.9565  & 0.9536  & 0.7970  & 0.7955  & 0.7500 \\
\midrule
\multicolumn{6}{c}{\textit{Pythia 410M}} \\
\midrule
Permutation & 0.3333  & 0.3953  & 0.3971  & 0.1564  & 0.1681  & 0.1368 \\
Orthogonal  & 0.8485  & 0.9286  & 0.9417  & 0.6667  & 0.5504  & 0.4763 \\
Invertible  & 0.9375  & 0.9439  & 0.9259  & 0.6556  & 0.5675  & 0.5152 \\
\midrule
\multicolumn{6}{c}{\textit{GPT2 XL 1.5B}} \\
\midrule
Permutation & 0.0000  & -0.0083  & 0.0000  & 0.0000  & -0.0185  & 0.0093 \\
Orthogonal  & 0.8462  & 0.7846  & 0.7521  & 0.4828  & 0.5294  & 0.5187 \\
Invertible  & 0.9452  & 0.9605  & 0.9697  & 0.7857  & 0.7119  & 0.6629 \\
\bottomrule
\end{tabular}%
}
\caption{Comparison of Top-K Syntax and Semantics summary metrics across permutation, orthogonal, and invertible transformations for Pythia 160M, Pythia 410M and GPT2-XL.}
\label{tab:language_baseline_comp}
\end{table}

\begin{figure}[h!]
\centering
\includegraphics[width=0.9\linewidth]{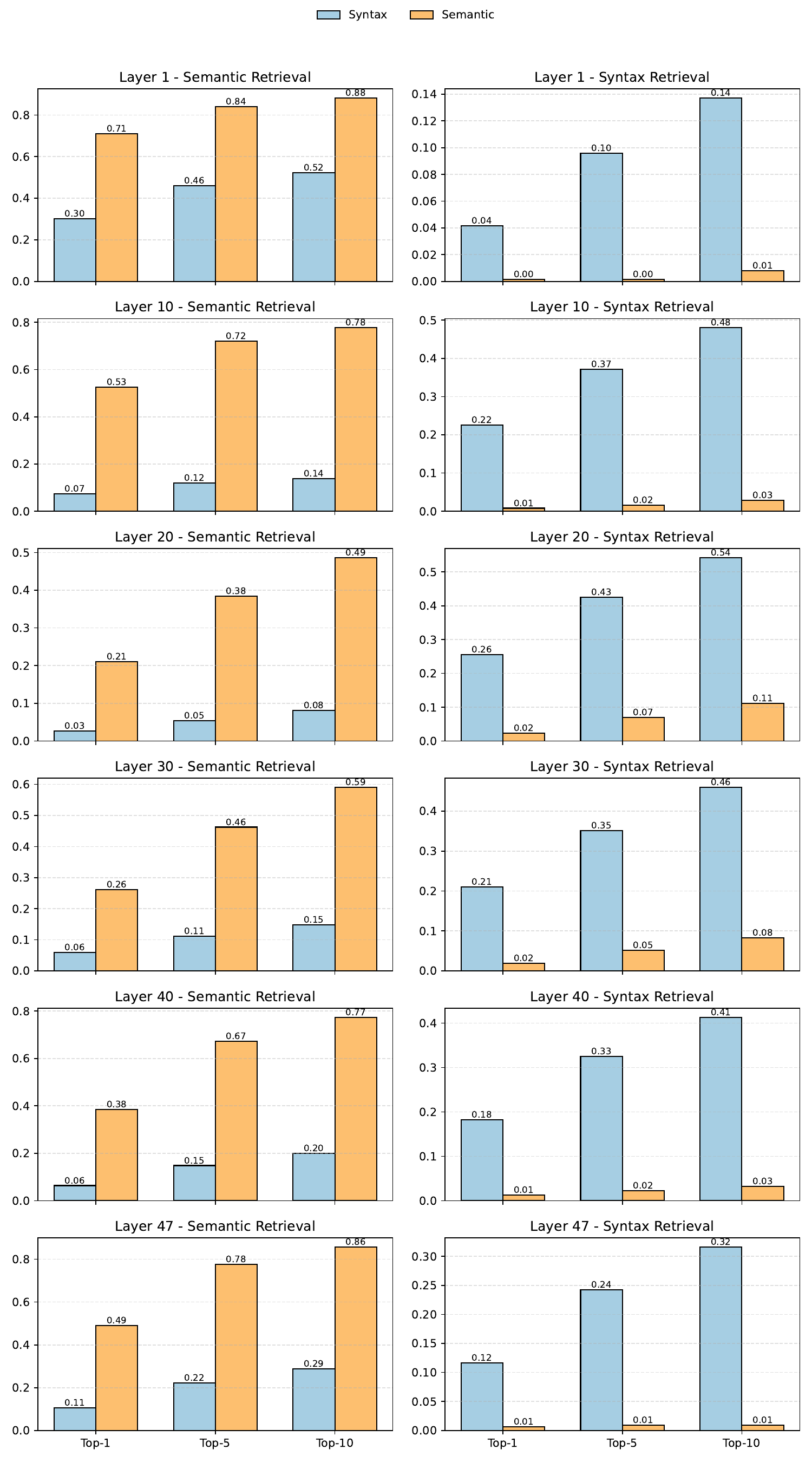}
\caption{(A) Top-K Semantic retrieval; (B) Top-K Syntax retrieval across different layers of GPT2-XL obtained using invertible transforms}
\label{fig:all_layers_comp}
\end{figure}

\clearpage

\subsection{Licenses and terms of use for existing assets}
\label{license}

Our experiments use publicly available datasets and pretrained model checkpoints obtained from their official release sites or repositories. Table~\ref{tab:asset_licenses} summarizes the corresponding licenses or access terms. For assets governed by access agreements or dataset-specific terms rather than standard software licenses, we report the official terms of use instead.
\begin{table}[t]
\centering
\footnotesize
\caption{Licenses and terms of use for existing assets used in this work.}
\label{tab:asset_licenses}
\begin{tabularx}{\columnwidth}{@{}l l X@{}}
\toprule
\textbf{Asset} & \textbf{Type} & \textbf{License / terms of use} \\
\midrule
ImageNet & Dataset & Official ImageNet access agreement / terms of use (not a standard open license). \\
THINGS & Dataset & Original THINGS images are for academic use only; the official site recommends using THINGSplus license-free images for publications. \\ 
MS-COCO & Dataset & Official COCO Terms of Use; images are subject to Flickr terms of use, and COCO annotations / website are distributed under CC BY 4.0. \\
Flickr8k-Expert & Dataset & Available for non-commercial academic research (images subject to original Flickr8k academic request terms). \\
NSD & Dataset & Access subject to the NSD Data Access Agreement. \\
Torchvision pretrained vision models & Model & BSD-3-Clause (official torchvision repository; includes standard pretrained models such as ResNet, VGG, AlexNet, and torchvision vision transformers where used). \\
timm pretrained vision models & Model & Official timm repository is Apache License 2.0; includes pretrained ConvNeXt / ViT / Swin models where used. \\
DINOv2 checkpoints & Model & Apache License 2.0. \\
OpenLLaMA & Model & Apache License 2.0. \\
Qwen2.5 & Model & Checkpoint-specific official Hugging Face license files; verified checkpoints used here include Apache License 2.0. \\
Qwen3 & Model & Apache License 2.0. \\
Gemma 2 & Model & Access subject to Google's Gemma usage license / access agreement. \\
ALBERT & Model & Apache License 2.0. \\
GPT-2 & Model & Hugging Face checkpoint license varies by released variant; verified pages show MIT for openai-community/gpt2 and Modified MIT License for gpt2-medium. \\
CLIP-ViT-B/32 & Model & MIT license. \\
Pereira 2018 & Dataset & NO license as mentioned in the official OSF website \\
\bottomrule
\end{tabularx}
\end{table}

\subsection{Impact Statement} 
\label{subsec:impact}

This work is primarily intended as an analytical framework for investigating the presence of modularity in high-dimensional representations learned by artificial models. In particular, we examine whether modular structure, even when not explicitly present in the native representational space, can emerge through orthogonal and invertible transformations. Such structure may provide insights with potential implications for interpretability, model comparison, and computational neuroscience.

\clearpage
\section*{NeurIPS Paper Checklist}

\begin{enumerate}

\item {\bf Claims}
    \item[] Question: Do the main claims made in the abstract and introduction accurately reflect the paper's contributions and scope?
    \item[] Answer: \answerYes{} 
    \item[] Justification: This paper investigates the feasibility of achieving modularity in neural network representations. The central finding is that while dissociation may not be attainable at the level of individual units, it becomes achievable through invertible linear transformations that preserve the intrinsic geometry of the learned representations. The methodology is rigorously developed in Section \ref{subsec:methods}, Appendix section \ref{subsec:constraints}, and the core claims are systematically validated in Section \ref{subsec:results} and Appendix Section \ref{subsec:geirhos_silhoutte}, \ref{subsec:geirhos_conflicting_cues}, \ref{subsec:Mante task}.

\item {\bf Limitations}
    \item[] Question: Does the paper discuss the limitations of the work performed by the authors?
    \item[] Answer: \answerYes{} 
    \item[] Justification: The discussion section explicitly discusses the limitations of the current paper, and scope for future work, such as required analysis of the functionality of the residual units and the optimal number of units required for modularity amongst others.

\item {\bf Theory assumptions and proofs}
    \item[] Question: For each theoretical result, does the paper provide the full set of assumptions and a complete (and correct) proof?
    \item[] Answer: \answerNA{} 
    \item[] Justification: The paper presents a methodological framework and formal definitions, but it does not state theorem-level theoretical results requiring formal proofs in the checklist sense. The mathematical content is used to define the contrastive training procedure rather than to prove new theoretical guarantees.

    \item {\bf Experimental result reproducibility}
    \item[] Question: Does the paper fully disclose all the information needed to reproduce the main experimental results of the paper to the extent that it affects the main claims and/or conclusions of the paper (regardless of whether the code and data are provided or not)?
    \item[] Answer: \answerYes{} 
    \item[] Justification: The paper provides substantial information needed to reproduce the main experiments, including the contrastive training procedure to obtain dissociations for vision and language model representations and the task sufficiency objective for recurrent neural network representations (section \ref{subsec:methods}, \ref{subsec:constraints}, \ref{app:rnnnetworkappendix}, , dataset descriptions in Appendix Section \ref{subsec:datasets}, model and pool specifications, and experimental setup / implementation details in Appendix Section \ref{subsec:model_details}. The main results are also tied to concrete datasets, pool constructions, and evaluation metrics in results Section \ref{subsec:results}. 

\item {\bf Open access to data and code}
    \item[] Question: Does the paper provide open access to the data and code, with sufficient instructions to faithfully reproduce the main experimental results, as described in supplemental material?
    \item[] Answer: \answerNo{} 
    \item[] Justification: The current submission does not yet include a public code release or complete reproduction package. We plan to release code and detailed reproduction instructions in the camera-ready version.

\item {\bf Experimental setting/details}
    \item[] Question: Does the paper specify all the training and test details (e.g., data splits, hyperparameters, how they were chosen, type of optimizer) necessary to understand the results?
    \item[] Answer: \answerYes{} 
    \item[] Justification: The paper provides detailed descriptions of all methodological components, including the network architectures, optimization procedures, learning rates, hardware specifications,  datasets and models used in our experiments in Sections \ref{subsec:methods}, \ref{subsec:datasets}, \ref{subsec:modeldetails}, \ref{subsec:exptdetails}, \ref{app:rnnnetworkappendix}.

\item {\bf Experiment statistical significance}
    \item[] Question: Does the paper report error bars suitably and correctly defined or other appropriate information about the statistical significance of the experiments?
    \item[] Answer: \answerYes{} 
    \item[] Justification: We have added error bars for all of our plots in section \ref{subsec:brain} and \ref{sec:rnn_results}. For sections 1,2, since experiments were done over a single split, error bars cannot be provided at the time.

\item {\bf Experiments compute resources}
    \item[] Question: For each experiment, does the paper provide sufficient information on the computer resources (type of compute workers, memory, time of execution) needed to reproduce the experiments?
    \item[] Answer: \answerYes{} 
    \item[] Justification: The paper provides detailed descriptions of all methodological components, including the network architectures, optimization procedures, learning rates, hardware specifications,  datasets and models used in our experiments in Sections \ref{subsec:methods}, \ref{subsec:datasets}, \ref{subsec:modeldetails}, \ref{subsec:exptdetails}, \ref{app:rnnnetworkappendix}.
    
\item {\bf Code of ethics}
    \item[] Question: Does the research conducted in the paper conform, in every respect, with the NeurIPS Code of Ethics \url{https://neurips.cc/public/EthicsGuidelines}?
    \item[] Answer: \answerYes{} 
    \item[] Justification: To the best of our knowledge, the research described in the paper conforms to the NeurIPS Code of Ethics. The work is a methodological and empirical investigation into the existence of modularity in ANN representations, and the submission preserves anonymity appropriately for review.

\item {\bf Broader impacts}
    \item[] Question: Does the paper discuss both potential positive societal impacts and negative societal impacts of the work performed?
    \item[] Answer: \answerYes{} 
    \item[] Justification: The paper includes an Impact Statement at the end in Appendix section \ref{subsec:impact} that addresses both positive and negative societal impacts. It identifies potential benefits for interpretability, model comparison, and computational neuroscience.
    
\item {\bf Safeguards}
    \item[] Question: Does the paper describe safeguards that have been put in place for responsible release of data or models that have a high risk for misuse (e.g., pre-trained language models, image generators, or scraped datasets)?
    \item[] Answer: \answerNA{} 
    \item[] Justification: The work analyzes existing pretrained models and public datasets. The paper does not release a new high-risk model, scraped dataset, or generative system that would require a safeguards discussion of the kind described in the checklist.

\item {\bf Licenses for existing assets}
    \item[] Question: Are the creators or original owners of assets (e.g., code, data, models), used in the paper, properly credited and are the license and terms of use explicitly mentioned and properly respected?
    \item[] Answer: \answerYes{} 
    \item[] Justification: Appendix Section 10, Table~\ref{tab:asset_licenses} includes a dedicated subsection and table summarizing the licenses or official access terms for the primary datasets and pretrained model assets used in the paper, including cases governed by data-access agreements or usage terms rather than a standard open-source license.

\item {\bf New assets}
    \item[] Question: Are new assets introduced in the paper well documented and is the documentation provided alongside the assets?
    \item[] Answer: \answerNA{} 
    \item[] Justification: The paper does not currently introduce or release a new dataset, model, or documented software asset as part of the submission. The contribution is a framework and empirical analysis built on existing assets, so this item is not applicable in the current form. 

\item {\bf Crowdsourcing and research with human subjects}
    \item[] Question: For crowdsourcing experiments and research with human subjects, does the paper include the full text of instructions given to participants and screenshots, if applicable, as well as details about compensation (if any)? 
    \item[] Answer: \answerNA{} 
    \item[] Justification: The paper does not report a new crowdsourcing study or new human-subject data collection performed by the authors. Instead, it uses previously collected public datasets with human annotations or neural recordings{}

\item {\bf Institutional review board (IRB) approvals or equivalent for research with human subjects}
    \item[] Question: Does the paper describe potential risks incurred by study participants, whether such risks were disclosed to the subjects, and whether Institutional Review Board (IRB) approvals (or an equivalent approval/review based on the requirements of your country or institution) were obtained?
    \item[] Answer: \answerNA{} 
    \item[] Justification:The paper does not involve new human-subject experimentation by the authors; it analyzes previously collected public datasets.{}

\item {\bf Declaration of LLM usage}
    \item[] Question: Does the paper describe the usage of LLMs if it is an important, original, or non-standard component of the core methods in this research? Note that if the LLM is used only for writing, editing, or formatting purposes and does \emph{not} impact the core methodology, scientific rigor, or originality of the research, declaration is not required.
    \item[] Answer: \answerNA{} 
    \item[] Justification: Language models are studied as experimental objects in this paper, but were not used as a non-standard component of the methodology. Any LLM use was limited to writing, editing, and formatting support.

\end{enumerate}

\end{document}